\documentclass{ieeeaccess}
  \usepackage{cite}
\usepackage{amsmath,amssymb,amsfonts}
\usepackage{algorithmic}
\usepackage{graphicx}
\graphicspath{ {./images/} }
\usepackage{CJK}
\usepackage{textcomp}
\usepackage{comment}
\usepackage{flushend}
\usepackage{xcolor, soul}
\sethlcolor{white}
\sethlcolor{red}
\usepackage{array}
\usepackage{multirow}
\usepackage{mathabx}
\usepackage{xcolor}
\usepackage{algorithm}
\usepackage{subcaption}
\usepackage[spaces,hyphens]{xurl}
\usepackage[colorlinks,allcolors=blue]{hyperref}

\newcommand\MyBox[2]{
 \fbox{\lower0.75cm
 \vbox to 1cm{\vfil
  \hbox to 1cm{\hfil\parbox{1.4cm}{#1\\#2}\hfil}
  \vfil}%
 }%
}

\def\BibTeX{{\rm B\kern-.05em{\sc i\kern-.025em b}\kern-.08em
 T\kern-.1667em\lower.7ex\hbox{E}\kern-.125emX}}
\begin{document}
\history{Date of publication xxxx 00, 0000, date of current version xxxx 00, 0000.}
\doi{10.1109/ACCESS.2022.0092316}

\title{Early Diagnosis and Severity Assessment of Weligama Coconut Leaf Wilt Disease and Coconut Caterpillar Infestation using Deep Learning-based Image Processing Techniques}

\author{\uppercase{Samitha Vidhanaarachchi}\authorrefmark{1,*},\IEEEmembership{Member, IEEE},
\uppercase{Janaka L. Wijekoon}\authorrefmark{2,3,*},\IEEEmembership{Senior Member, IEEE}, \uppercase{W. A. Shanaka P. Abeysiriwardhana}\authorrefmark{4}, \IEEEmembership{Member, IEEE}, \uppercase{and Malitha Wijesundara}\authorrefmark{1},\IEEEmembership{Member, IEEE}
}

\address[1]{Sri Lanka Institute of Information Technology, New Kandy Rd, Malabe 10115, Sri Lanka}
\address [2]{Victorian Institute of Technology, Adelaide Campus, South Australia}
\address[3]{Department of System Design Engineering, Keio University, Yokohama, Japan}
\address[4]{ACSL Ltd 3-6-4 Rinkaicho Edogawa, Tokyo 134-0086 Japan}

\markboth
{Author \headeretal: Preparation of Papers for IEEE TRANSACTIONS and JOURNALS}
{Author \headeretal: Preparation of Papers for IEEE TRANSACTIONS and JOURNALS}

\corresp{Corresponding author: Samitha Vidhanaarachchi (e-mail: samithapva@gmail.com).}

\begin{abstract}
Global Coconut (Cocos nucifera (L.)) cultivation faces significant challenges, including yield loss, due to pest and disease outbreaks. In particular, Weligama Coconut Leaf Wilt Disease (WCWLD) and Coconut Caterpillar Infestation (CCI) damage coconut trees, causing severe coconut production loss in Sri Lanka and nearby coconut-producing countries. Currently, both WCWLD and CCI are detected through on-field human observations, a process that is not only time-consuming but also limits the early detection of infections. This paper presents a study conducted in Sri Lanka, demonstrating the effectiveness of employing transfer learning-based Convolutional Neural Network (CNN) and Mask Region-based-CNN (Mask R-CNN) to identify WCWLD and CCI at their early stages and to assess disease progression. Further, this paper presents the use of the You Only Look Once (YOLO) object detection model to count the number of caterpillars distributed on leaves with CCI. The introduced methods were tested and validated using datasets collected from Matara, Puttalam, and Makandura, Sri Lanka. The results show that the proposed methods identify WCWLD and CCI with an accuracy of 90\% and 95\%, respectively. In addition, the proposed WCWLD disease severity identification method classifies the severity with an accuracy of 97\%. Furthermore, the accuracies of the object detection models for calculating the number of caterpillars in the leaflets were: YOLOv5-96.87\%, YOLOv8-96.1\%, and YOLO11-95.9\%.
\end{abstract}


\titlepgskip=-15pt

\maketitle

\section{Introduction}
\label{sec:introduction}
The Cocos nucifera (L.) (commonly known as coconut) tree is a member of the palm family, which has a 50-60-year economic lifespan and starts bearing fruits within 5-6 years after being planted. Usually, the tree grows up to 30 m with a crown of 30-40 leaves, making an ample amount of food and shelter for the survival and reproduction of many insects and pests \cite{Kumara2015StatusAM}. Pests and diseases in coconut plants cause fatal damage to the palm and can result in outbreaks, causing heavy economic losses \cite{Kumara2015StatusAM}. Therefore, efficient disease and pest-controlling methods are vital for coconut cultivation \cite{Chandy2019}.

Recent studies \cite{inproceedings,L.C.P.Fernando,L.C.P.Fernando2, L.C.P.Fernando3, coconutcaterpillar,coconutcaterpillar2} discuss that Weligama Coconut Leaf Wilt Disease (WCLWD) and Coconut Caterpillar Infestation (CCI) pose the greatest threats to coconut plantations in Sri Lanka and regional countries, causing rapid and severe damage to the coconut cultivation. According to \cite{L.C.P.Fernando,L.C.P.Fernando2, L.C.P.Fernando3}, WCLWD, which is an incurable disease, has been detected to be spreading at an alarming rate in Sri Lanka due to lack of early detection capabilities. During the expert survey, it was confirmed by principal entomologists from the Coconut Research Institute of Sri Lanka (CRISL) \cite{CRISL} WCLWD symptoms, specifically uneven yellowing \cite{ WCLWDyellowing, inproceedings, Nainanayaka}, are frequently misclassified as the yellowing caused by Magnesium (Mg) deficiency \cite{WCLWDyellowing, Nainanayaka}, making it difficult for even trained individuals to differentiate the two \cite{inproceedings}. Similarly, the dried appearance of leaves caused by CCI is also difficult to distinguish from Leaf Scorch Decline (LSD) \cite{inproceedings}, which is a physiological disorder. For these reasons, the majority of coconut growers are incapable of identifying the WCLWD and CCI infestations at their early stages.

The principal entomologists also confirmed that both WCLD and CCI are detected using on-field human observations, which is not only time-consuming but also limits the early detection of infections. ICT tools have been successfully incorporated into developing pest and disease management practices in the recent past to early identification and treatments \cite{dharmadhikari1977short, Vidhanaarachchi}. Also, recent literature has shown that Machine Learning (ML) and artificial intelligence (AI) are thriving in agricultural research studies, significantly improving precision farming through automated analysis and decision-making \cite{{presiAgri1,presiAgri2}}. Among these advancements, the use of computer vision and deep learning (DL) techniques, such as feature extraction to automate pest and disease classification, is gaining momentum as a valuable tool to modernize conventional observation methods \cite{Kumara2015StatusAM, Elison, Chandy2019, HanKadipa, YangYu, Francl199757, Federico}. Consequently, as an extension of the preliminary study \cite{Vidhanaarachchi}, this paper presents a study on the effectiveness of using DL techniques for the early identification of CCI, automating the detection of caterpillars, and the early identification of WCLWD.

\section{Background Survey}
\label{sec:Background}
\begin{table*}[t]
\centering
\caption{Summary of significant ML and DL applications in agriculture}
\label{tab:AI_agriculture}
\begin{tabular}{l|l|l|l|l|l} 
\hline
\multicolumn{1}{c|}{\textbf{Ref}} &
 \multicolumn{1}{c|}{\textbf{Crop}} &
 \multicolumn{1}{c|}{\textbf{Objectives}} &
 \multicolumn{1}{c|}{\textbf{Methods}} &
 \multicolumn{1}{c|}{\textbf{Respective Accuracy's}} &
 \multicolumn{1}{c}{\textbf{Year}} \\ \hline
\cite{Sladojevic} &
 Pear, Cherry etc... &
 Disease classification &
 CNN-based models &
 96.3\% &
 2016 \\
\cite{HanKadipa} &
 Black Gram &
 Nutrient deficiency classification &
 Transfer learning-based CNN &
 65.44\% &
 2019 \\
\cite{Francl199757} &
 Wheat &
 Disease propagation prediction &
 ANN, Logistic Models &
 93\% \& 90\% &
 1997 \\
\cite{Wijekoon} &
 Tomato, Beans etc... &
 Disease detection \& dispersion &
 CNN, Mask R-CNN, Gaussian Plume &
 90\% - 94\% &
 2022 \\
\cite{Federico} &
 Tomato &
 Disease detection &
 Near-infrared spectroscopy &
 78\% &
 2004 \\
\cite{Huang} &
 Moth Orchids &
 Disease detection &
 Gray level co-occurrence matrix &
 89.6\% &
 2007 \\
\cite{Wang} &
 Tomato &
 Disease detection &
 Faster R-CNN, Mask R-CNN &
 88.53\% \& 99.64\% &
 2021 \\
\cite{MaskRCNNGeetha} &
 Cabbage &
 Pest Detection &
 CNN, Mask R-CNN &
 95.3\% \& 98.3\% &
 2024 \\
\cite{Manoharan} &
 Guava, Citrus etc... &
 Nutrition deficiency classification &
 CNN-based model &
 88\% &
 2020 \\
\cite{Hewawitharana} &
 Tea &
 Disease severity assessment &
 CNN, YOLOv8, Mask R-CNN &
 89.9\%, 98\% \& 89\% &
 2023 \\
\cite{Kasinathan2023} &
 Maize &
 Pest Detection &
 Mask RCNN &
 94.21\% &
 2023 \\
\cite{Elison} &
 Wheat, Maize etc... &
 Pest classification \& counting &
 OpenCV, ANN &
 97.9\% \& 92.5\% &
 2020 \\
\cite{Chandy2019} &
 Coconut &
 Pest and disease identification &
 DNN &
 Only Proposed &
 2019 \\
\cite{MARAY2022108399} &
 Coconut &
 Disease Classification &
 Gated Recurrent Unit &
 97.75\% &
 2022 \\
\cite{Kadethankar} &
 Coconut &
 Infestation Classification &
 CNN, Faster R-CNN &
 84.64\% \& 97.30\% &
 2021 \\
\cite{SINGH2021105986} &
 Coconut &
 Disease Classification &
 CNN &
 96.94\% &
 2021 \\ \hline
\end{tabular}
\end{table*}

This section explores the application of DL and Machine Learning (ML) in agriculture, focusing on disease and pest management. The section covers the use of DL models, CNNs, instance segmentation, and object detection techniques in identifying and classifying plant diseases, nutrient deficiencies, and pests. The review also highlights specific studies and their findings in these areas, emphasizing the potential and current limitations in the field. A comprehensive summary is presented in Table \ref{tab:AI_agriculture}.

Several studies have explored the use of DL techniques for disease identification and management. For instance, Sladojevic et al. \cite{Sladojevic} and Han et al. \cite{HanKadipa} have highlighted the success of CNN-based models in classifying healthy and diseased leaves, as well as detecting nutrient deficiencies in black gram. The work of Francl et al. \cite{Francl199757}, further underscores the application of Artificial Neural Network (ANN) in predicting disease appearances and identifying specific plant diseases with high precision and recall rates. Additionally, studies by Miriyagalla et al. \cite{Miriyagalla}, Wijekoon et al. \cite{Wijekoon}, Hahn et al. \cite{Federico}, and Huang \cite{Huang} have contributed to the development of platforms for predicting disease spread and detecting specific plant diseases using advanced technologies like near-infrared spectroscopy and deep neural networks. Even though such studies justify that the use of DL technologies together with image processing is very effective, none of those studies focused on the coconut industry, making it a niche sector to apply such advanced techniques for the early identification of diseases.

Instance segmentation and object detection technologies have also become prominent tools for pest and disease detection in plants. Wang et al. \cite{Wang} detailed the utilization of Faster R-CNN and Mask R-CNN to identify and segment plant diseases accurately. Geetha et al. \cite{MaskRCNNGeetha} advanced this field by addressing the challenges of detecting small pests in natural environments and classifying them into multiple categories. Their methodologies have significantly enhanced the accuracy and efficiency of in-field pest detection systems, which are crucial for sustainable crop management. Manoharan et al. \cite{Manoharan} employed Mask R-CNN to calculate the extent of nutritional deficiencies (Nitrogen, Phosphorous, and Potassium) in crops like Guava, Groundnut, and Citrus by assessing the degree of leaf unhealthiness through masked images. Hewawitharana et al. \cite{Hewawitharana} addressed the persistent threat of Blister Blight (BB) in the Ceylon tea industry and provided classification and severity assessment using CNNs and Mask R-CNN. Additionally, Tiwari et al. in \cite{Kasinathan2023} reported the detection of Fall Armyworm (FAW) insects in field crops using Mask R-CNN, leading to the creation of bounding boxes and segmentation frames for each insect in the image. Unfortunately, none of these studies were focused on coconuts. 

The automation of pest management has also been a focus in recent literature, as evidenced by the work of Lins et al. \cite{Elison}, who developed software for automated counting and classifying aphids. This method addresses the inefficiencies and errors associated with manual counting. However, there remains a gap in the automation of counting coconut caterpillars, as they still resort to manual counts in current practices.

It was identified that there is  limited literature specifically focused on coconut diseases and pest. Among them, Chandy’s study \cite{Chandy2019} proposed a DL-based system identifying coconut pest infestations, but it did not extend to disease classification. However, several studies have employed DL techniques for classifying diseases in coconut trees. Maray et al. \cite{MARAY2022108399} introduced the AIE-CTDDC model for disease detection, but not for severity assessment. Kadethankar et al. \cite{Kadethankar} presented an end-to-end pipeline for detecting rhinoceros beetle infestations in coconut trees using drone imagery. Their approach focused on detecting and extracting individual tree crowns from drone images, enabling targeted analysis. Singh et al. \cite{SINGH2021105986} conducted disease classification using CNN, but they identified severity assessment and DL segmentation methods as future research direction. These are also two main objectives of the present study.

As summarized in Table \ref{tab:AI_agriculture}, there is a clear need for further research to leverage advanced AI techniques for improved pest and disease management in coconut cultivation because as stated perviously, research on coconut pest and disease management is relatively limited, with most studies focusing on disease classification rather than severity assessment or advanced detection techniques.

Therefore, this study was conducted with three main objectives: 

\begin{enumerate}
\item Analyze the effectiveness of using DL techniques to identify and classify WCLWD and CCI at their early stages
\item Calculate the degree of diseased conditions of WCLWD using transfer learning-based CNN
\item Measure the progression of CCI using advanced deep learning segmentation techniques such as Mask R-CNN and You Only Look Once (YOLO) object detection models, helping farmers determine the necessary precautionary management actions.
\end{enumerate}

The experiments were conducted in hot spot areas in the Matara and Puttalam districts of Sri Lanka, and the datasets were collected from different locations and times to reduce class imbalances; the dataset is published in \cite{kaggle}.

\section{Materials and methods}
\label{sec:Materials}

\subsection{Data collection}
\label{sec:DataCollection}

To develop the ML models for early detection and classification of CCI and WCLWD, we first collected images across a range of conditions, including the following:

\begin{itemize}
\item Healthy and disease-infected trees of different diseases, including flattened leaflets, uneven yellowing, tip browning, CCI, and WCLWD.
\item Tree fonds of different sizes and growth.
\item Different locations in the country.
\item Different stages of the growth of both diseases.
\item Different times of the day(morning and evening).
\item Diverse weather conditions.
\item Different image qualities and resolutions. 
\end{itemize}

It should be noted that the two conditions below were applied to collect data from disease-infected leaves, as we were informed by field experts from the CRISL.  
\begin{itemize}
\item For WCLWD-infected leaves, images were collected from the upper leaf surface.
\item For CCI-infected leaves, leaflet images were acquired from the lower surface.
\end{itemize}
Images were captured using mobile phones and Digital Single Lens Reflect (DSLR) cameras. To cover a wider range of sensor and image qualities, iPhone6 with 8MP camera \cite{iphone6}, iPhone11 with 12MP camera \cite{iphone11}, and a high-quality Canon ESO 3000D DSLR with 18MP camera with a larger APS-C sensor \cite{canon} were used for data collection. Note that auto settings with auto-post processed JPEG output formatting were used in both scenarios. These conditions were aimed at aligning the data collection process with the traditional, manual assessments of diseases that have been practised within the CRISL \cite{L.C.P.Fernando}.

The acquisition of leaf samples was carried out in collaboration with research personnel affiliated with the CRISL \cite{CRISL}, within the time frame of April to May 2021. The sample images were classified and verified by skilled personnel from the CRISL, and this was considered the ground truth for this study. This approach was undertaken to mitigate potential similarities between image samples, thus ensuring a no class imbalance in the dataset. 

A summary of the acquired dataset is given in Table \ref{tab:SummaryofDataSamples}.
The WCLWD dataset employed for disease classification comprised a total of 9,258 images, while 3,307 images were used to evaluate symptom severity. The CCI infestation classification task was performed using a dataset consisting of 1,600 images, while the ML model for quantifying the number of caterpillars on a leaf was trained using 1,400 images. As one of the contributions of this study, we made the curated dataset available for researchers in this domain via Kaggle \cite{kaggle}. 

\begin{table*}[t]
\caption{Summary of Data Samples}
\label{tab:SummaryofDataSamples}
\renewcommand{\arraystretch}{1.5}  
\centering
\begin{tabular}{lllc|ccc}
\hline
\textbf{Condition} & \textbf{Location} & \textbf{Purpose} & 
& \textbf{Training} & \textbf{Validation} & \textbf{Testing (collected separately)} \\ \hline
\multirow{2}{*}{WCLWD} & \multirow{2}{*}{Weligama, Matara} & Disease Classification & 
& 7406 & 1852 & 117 \\
            &                  & Symptom Severity    & 
& 2645 & 662 & 106 \\ 
\multirow{2}{*}{CCI}  & \multirow{2}{*}{Puttalam, Lunuwila} & Infestation Classification &  
& 1280 & 320 & 100 \\
            &                   & Caterpillar Calculation  &  
& 1120 & 280 & 140 \\ \hline
\end{tabular}
\end{table*}

The coconut fronds infected by WCLWD were sourced from the city of Matara, Sri Lanka, an area known for disease prevalence. Matara is located within the low country wet agro-ecological zone of the southern province of Sri Lanka (latitude 6°-10° and longitude 79°-82°). According to research \cite{kumara2015prevalence}, WCLWD is isolated to these areas, appearing in pockets or clusters. Extensive containment measures, such as the removal and burning of infected palms, have been implemented. Hence, this area was selected for data collection, with the few remaining affected areas kept under strict observation and made accessible only for research purposes, allowing data to be gathered under carefully controlled conditions. Fronds exhibiting symptoms such as flattened leaflets, uneven yellowing, and tip browning were selected for image collection, as they are the initial symptoms of WCLWD \cite{inproceedings}.

For CCI, experimental images were obtained from multiple estates in Lunuwila and Makandura, Sri Lanka. Both these estates are situated within the low-country dry zone of the Puttalam district, a part of the area known for significant coconut cultivation, commonly referred to as the Coconut Triangle \cite{coconut_triangle} \footnote{This geographical triangle includes Puttalam, Kurunegala, and Gampaha districts, which have excellent climatic conditions for coconut palms.}.   Caterpillar damage is prevalent within the Coconut Triangle, thus this area was considered for data collection. We were careful to select leaflets displaying dried brown patches with caterpillar galleries for the dataset. 

\subsection{Data Preprocessing}
The collected raw data underwent a series of preprocessing steps to enhance accuracy and to reduce calculation complexity. Image sizes ranging from 150 × 150 to 750 × 750 dimensions were collected during the field visit. To provide consistent input dimensions for neural networks, all images were resized to a uniform size of 300 × 300 pixels. Further, the pixel values of the images were normalized by dividing them by 255, scaling them to a range between 0 and 1. Subsequently, data augmentation techniques of rotation, filling, shearing (both horizontally and vertically), flipping (both horizontally and vertically), and zooming were applied to expand the sample size and mitigate the potential of model overfitting. The deep learning models were trained using pre-processed coconut leaflet images, with the dataset split into an 80\% training set and a 20\% validation set. The test data were then collected separately to avoid manual pre-selection and biasness during field visits.

\subsection{Data Annotation}
Initially, all preprocessed and augmented images were manually annotated by skilled professionals at CRISL. VGG Image Annotator (version 2.0.11) \cite{VIA_software} was utilized by the CRISL experts to leverage their specialized domain knowledge for accurately identifying and labelling disease symptoms. This tool was selected owing to the effectiveness and the accuracy presented in agricultural research \cite{vggref1,vggref2}. The polygon annotation labelling technique was then used to mark the area of the leaflets. During the process, leaf images with and without infestation damage were labelled, and the remaining region was set to the background, and the output file contains the coordinates for the polygonal regions. To train the model, these defined regions were fed into the Mask R-CNN model as input neurons.

It is important to note that the images used to train the YOLOv5 \cite{Yolov5}, YOLOv8 \cite{autogyro_yolo_v8_2024} and YOLO11 \cite{ultralytics2024} models were annotated using MakeSense.ai\cite{makesense_ai}, an open-source tool widely adopted in several similar studies, such as \cite{makesenceref1} and \cite{makesenceref2}, for creating precise annotations in the field of plant disease detection. The labeling technique employed was box annotation. MakeSense.ai also generates a single JSON file containing the coordinates of the annotated box regions and associated metadata. These defined regions were subsequently used as input data for the YOLO models during training.

\subsection{Classify WCLWD and assess severity level} 

WCWLD can be distinguished using four key symptoms as listed below:
\begin{enumerate}
  \item Flaccidity \footnote{Flaccidity is the initial symptom that aids in detecting infected palms during the early stages.}
  \item Uneven yellowing
  \item Leaf tip browning
  \item Breaking off the leaf tips
\end{enumerate}

\begin{figure*}[!t]
 \centering
 \begin{subfigure}[t]{0.45\linewidth}
  \includegraphics[width=\linewidth]{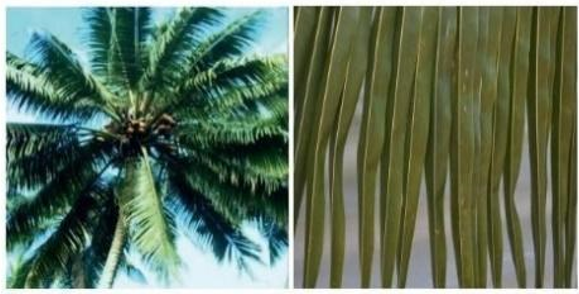}
  \subcaption{Flaccidity}
 \end{subfigure}
 \hfill
 \begin{subfigure}[t]{0.45\linewidth}
  \includegraphics[width=\linewidth]{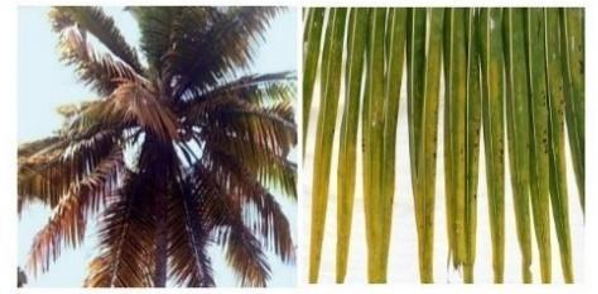}
  \subcaption{Uneven yellowing}
 \end{subfigure}
 \vskip\baselineskip
 \begin{subfigure}[t]{0.45\linewidth}
  \includegraphics[width=\linewidth]{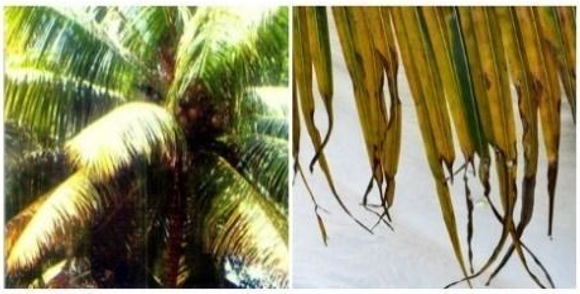}
  \subcaption{Drying of the leaflets}
 \end{subfigure}
 \hfill
 \begin{subfigure}[t]{0.45\linewidth}
  \includegraphics[width=\linewidth]{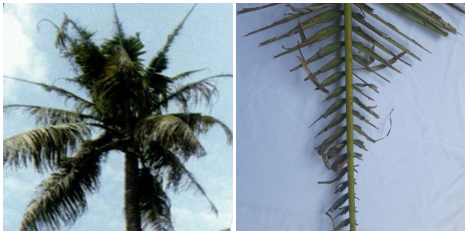}
  \subcaption{Breaking off the leaf tips}
 \end{subfigure}
 \caption{Symptoms of WCLWD. Images include (a) Flaccidity, (b) Uneven yellowing, (c) Drying of the leaflets, and (d) Breaking off the leaf tips.}
 \label{fig:WLWDstages}
\end{figure*}

The fourth symptom, breaking off the leaf tips, is distinguishable, and at this stage, there is no value in the coconut tree. Therefore, as one of the main scientific contributions of this study is to detect the disease at an early stage, the first three symptoms, which are common in the early phases, were chosen as the most important \cite{inproceedings}. The symptoms of WCWLD are shown in Fig. \ref{fig:WLWDstages}.

In this study, WCWLD was initially classified, and then disease severity was identified based on the progression of symptoms. The disease classification was performed using the flaccidity of the leaves, as it is the first symptom to appear, as shown in Fig. \ref{fig:WLWDstages}. This binary classification distinguished between healthy leaves and those exhibiting flaccidity. After classifying the disease based on flaccidity, disease severity was assessed using the remaining two symptoms: uneven yellowing and tip browning, which appear a few weeks later as the disease advances. To assess severity, a multi-class classification model with three classes—flaccidity, uneven yellowing, and tip browning—was trained. This strategy led us to identify and classify the disease at its early stages.

We trained ResNet50, DenseNet121, ResNet50V2, InceptionResNetV2, InceptionV3, MobileNetV2, Xception, and VGG16 pre-trained CNN models in this study using the sample data provided in Table \ref{tab:SummaryofDataSamples}. These architectures were compared to determine the best fit for classifying the disease and its severity. The models were trained to predict the output of the CNN with binary classification to identify flaccidity and classify the disease. 

Similarly, a multi-class classification model featuring three classes was trained to evaluate the severity of symptoms. CNNs are highly effective in image classification due to their ability to learn hierarchical features from images through a series of convolutional and pooling operations. The convolution operation involves applying a filter to the input image to extract important features like edges and textures. This operation can be expressed using \eqref{eq:convolution}.

\begin{equation}
\label{eq:convolution}
(I * h)(x, y) = \sum_{i=-k}^{k} \sum_{j=-k}^{k} I(x+i, y+j) \cdot h(i, j)
\end{equation}

where \( I \) represents the input image matrix, with \( I(x, y) \) denoting the pixel value at coordinates \( (x, y) \), and \( h \) is the filter or kernel matrix applied to detect features within the image. The indices \( i \) and \( j \) represent the coordinates within the filter, ranging from \( -k \) to \( k \), where \( k \) defines the filter size. The notation \( \sum_{i=-k}^{k} \sum_{j=-k}^{k} \) signifies the summation over all values in the filter region, allowing the weighted sum of pixel values centered at each \( (x, y) \) location.

After the convolution operation, the Rectified Linear Unit (ReLU) activation function is applied. The ReLU function is defined using \eqref{eq:relu}

\begin{equation}
\label{eq:relu}
f(x) = \max(0, x)
\end{equation}

where \( f(x) \) represents the ReLU activation function, which outputs the maximum of 0 and function’s input variable \( x \).This ensures that all negative values in the feature maps are set to zero, retaining only the positive activations, which helps the network capture complex patterns relevant to disease detection. To further reduce the dimensionality of the feature maps, we applied max pooling \eqref{eq:maxpooling}, which selects the maximum value from a defined window.

\begin{equation}
\label{eq:maxpooling}
P = \max\{ x_{i,j} \} \quad \text{for} \quad i,j \in \text{window size}
\end{equation}

where \( P \) denotes the output of the max pooling operation, which selects the maximum value \( x_{i,j} \) within a specified window size. The indices \( i \) and \( j \) represent the coordinates within this window. Max pooling reduces the spatial dimensions of the input while retaining the most prominent features, thereby enhancing computational efficiency and controlling overfitting in deep learning models. In the final stages of the network, a fully connected layer combines the features extracted from the convolutional layers to form a classification decision. The fully connected layer (dense layer) computes the output using \eqref{eq:fullyconnected}

\begin{equation}
\label{eq:fullyconnected}
z = W^T x + b
\end{equation}

where \( z \) represents the output of a dense layer. In this equation, \( W \) is the weight matrix, \( x \) is the input vector, and \( b \) is the bias vector.

To optimize the performance of the CNN models, the Adam and Stochastic Gradient Descent (SGD) optimization algorithms were applied through hyperparameter tuning. This allowed the network to iteratively minimize the loss function, improving the accuracy of the disease classification. The symptoms in each image were analyzed to determine the stage of the disease. If all three symptoms—flaccidity, uneven yellowing, and tip browning—were present, it indicated that the disease had progressed to a severe stage. However, if only flaccidity was observed, it suggested that the disease was still in its early stages. This approach helped assess the severity and progression of the disease based on the visible symptoms in the images.

\subsection{Identify and classify CCI} 
\begin{figure*}[!t]
\centerline{\includegraphics[width=\linewidth]{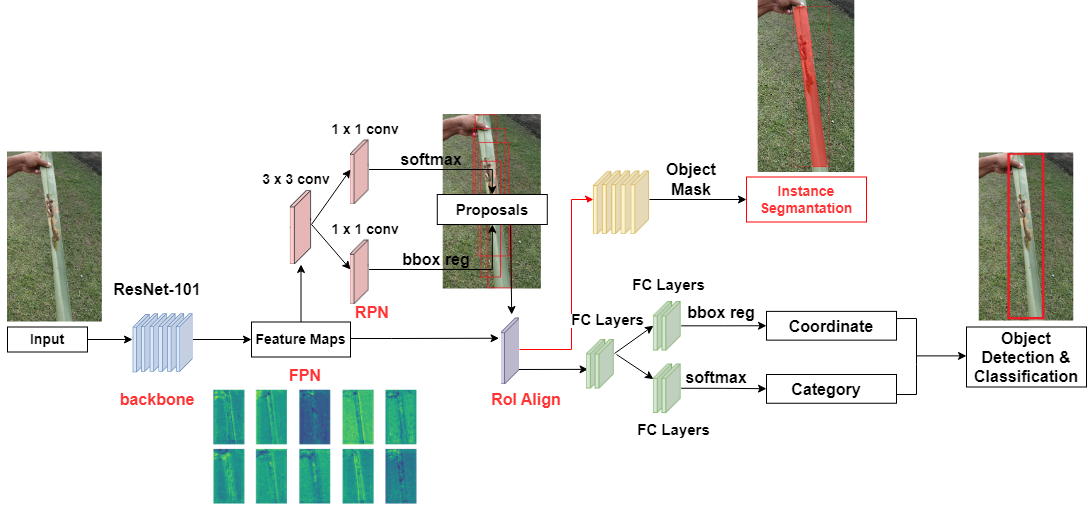}}
\caption{The training procedure of Mask R-CNN model for CCI instance segmentation}
\label{fig:ImgMaskRCNN}
\end{figure*}

Figure \ref{fig:ImgMaskRCNN} illustrates the proposed Mask R-CNN method, to detect CCI-infected areas and filter out those regions to classify the infested leaf. This process aids in segmenting the leaflet and in calculating disease progression. As depicted in the figure, first, a Region Proposal Network (RPN) was used to scan the feature maps to propose candidate bounding boxes that represent region of interest (RoIs) for further analysis. The RPN minimizes the objective function as expressed in \cite{he2017mask}.

Based on the studies conducted in \cite{YangYu} and \cite{Manoharan}, Resnet101 was used as the backbone CNN network for CCI classification due to its effectiveness compared to other available models. Resnet101 was combined with a Feature Pyramid Network (FPN) for feature extraction. As illustrated in Fig. \ref{fig:ImgMaskRCNN}, each stage of the ResNet101 architecture correlates to a different scale of feature maps. The feature pyramid of the FPN network is constructed using these feature maps.

The feature maps returned from the ResNet101 backbone were sent to the RPN to determine regions of interest (RoIs) where the leaflets exist. Intersection over Union (IoU) values \cite{Rahman} were used to mark the RoIs. As calculated using \eqref{eq:iou}, IoU is the ratio between the intersection and union of the ground truth bounding box and the model's bounding box. The ground truth bounding box and predicted bounding box are represented by \(A\) and \(B\), respectively. \(|A \cap B|\) is the area of overlap or the intersection, while \(|A \cup B|\) is the area of union. If the IoU of each predicted region is greater than or equal to 0.5, RoIs are considered. The rest are defined as the background (refer to Fig. \ref{fig:ImgMaskRCNN}).

\begin{equation}
\label{eq:iou}
\text{IoU} = \frac{|A \cap B|}{|A \cup B|}
\end{equation}

A bounding box rectangles were used to represent each RoI region acquired on the leaflet image, and Non-maximum suppression (NMS) \cite{Neubeck} was used to choose the bounding box with the highest foreground value when several bounding boxes overlap (when several bounding boxes represent one detection).Above process was followed by adjusting the dimensions of the anchor boxes with RoIAlign to provide a standard-sized output. Finally, the RoI features are used in the Fully Connected (FC) layer to conduct target classification of healthy and infected leaflets and bounding-box regression to mark the region.

\subsection{Calculate CCI progression level} 

The extent of infestation throughout the leaflet is referred to as the progression level. A pixel-level instance segmentation conducted using a Fully Convolutional Network (FCN) was utilized to mask the infested leaflets (refer to Fig. \ref{fig:ImgMaskRCNN}), in addition to the current classification and bounding box regression methods used to detect CCI. Instance segmentation was specifically considered here to differentiate each instance of CCI identification. This technique identifies each object’s instance by using a mask representation in the image, which simultaneously performs object class prediction and mask extraction providing distinct labels for separate occurrences of objects within the same class. 

Consequently, the leaf area of the image was masked, and only the class label defined as CCI was processed to calculate the disease progression level. Then, image crop segmentation was applied using the mask regions to extract the infected leaflet from the background using the algorithm \ref{alg:Alg1}, based on the method presented in the research paper \cite{cropsegmentation}. The result is an image where only the infected leaflets are visible, while the rest of the background is rendered black. 

\begin{algorithm}[!b]
\caption{Apply Crop Segmentation}
\label{alg:Alg1}
\begin{algorithmic}[1]
\STATE \textbf{Input:} RGB\_Leaf\_image, Masks
\STATE \textbf{Output:} Masked\_images

\STATE Initialize
\STATE $i \leftarrow 0$ \COMMENT{Mask index}
\STATE $n\_masks \leftarrow$ number of Masks
\STATE $Masked\_images \leftarrow$ empty array
\STATE $empty\_matrix \leftarrow$ size of RGB\_Leaf\_image

\WHILE{$i < n\_masks$}
  \STATE $Masked\_images.add(empty\_matrix)$
  \STATE $j \leftarrow 0$ \COMMENT{RGB channel index}
  \FOR{$j < 3 $}
    \STATE $j = j+1$
    \STATE $channel \leftarrow RGB\_Leaf\_image[:,:,j]$
    \STATE $Masked\_images[i][:,:,j] \leftarrow channel * mask[i]$
  \ENDFOR
  \STATE $i = i+1$
\ENDWHILE

\RETURN Masked\_images
\end{algorithmic}
\end{algorithm}

\begin{algorithm}[!t]
\caption{Color Segmentation, K-means Clustering, and Progression Calculation}
\label{alg:Alg2}
\begin{algorithmic}[1]
\STATE \textbf{Input:} (img) Crop Segmented Image
\STATE \textbf{Output:} Progression \%
\STATE k $\gets$ no\_of\_categories \text{ (Number of clusters/categories)}
\STATE hsv\_img $\gets$ \text{TO\_HSV(img)}
\STATE low\_g, up\_g $\gets$ \text{Green HSV thresholds}
\STATE low\_b, up\_b $\gets$ \text{Brown HSV thresholds}
\STATE g\_pixels $\gets$ \text{IN\_RANGE(hsv\_img, low\_g, up\_g)}
\STATE b\_pixels $\gets$ \text{IN\_RANGE(hsv\_img, low\_b, up\_b)}
\STATE img[g\_pixels] $\gets$ (20, 255, 10)
\STATE img[b\_pixels] $\gets$ (19, 69, 139)
\STATE img[!(g\_pixels \text{ OR } b\_pixels)] $\gets$ (255, 255, 255)
\STATE kmeans $\gets$ \text{INIT\_KMEANS(k)}
\STATE lbls $\gets$ \text{kmeans.labels\_}
\STATE counts $\gets$ \text{INIT\_COUNT\_ARRAY(k)}
\STATE \textbf{for each} lbl \textbf{in} lbls:
\STATE \ \ \ \ counts[lbl] $\gets$ counts[lbl] + 1
\STATE g\_color $\gets$ [20, 255, 10]
\STATE b\_color $\gets$ [19, 69, 139]
\STATE g\_count, b\_count $\gets$ 0
\STATE \textbf{for each} idx, count \textbf{in} counts:
\STATE \ \ \ \ \textbf{if} centers[idx] $\textbf{=}$ g\_color \textbf{then} g\_count $\gets$ count
\STATE \ \ \ \ \textbf{else if} centers[idx] $\textbf{=}$ b\_color \textbf{then} b\_count $\gets$ count
\STATE \ \ \ \ \textbf{end if}
\STATE \textbf{end for}
\STATE total\_count $\gets$ g\_count + b\_count
\STATE green\_perc $\gets$ \text{ROUND(g\_count / total\_count * 100)}
\STATE brown\_perc $\gets$ \text{ROUND(b\_count / total\_count * 100)}
\STATE \textbf{return} (green\_perc, brown\_perc)
\end{algorithmic}
\end{algorithm}

After crop segmentation, color segmentation was utilized to replace a range of HSV colors corresponding to green (healthy) and brown (necrotic leaf regions due to feeding of caterpillars) with single RGB pixel values ((20,255,10) \& (19,69,139) respectively). This approach addresses the issue of color variations in the image, where multiple shades of green and brown are present. The remaining (background) was turned white (255,255,255) to obtain a clear image. Therefore, after color segmentation, only three clusters of colors remained in the image: green, brown, and white. According to experts at CRISL, these are the three categories that should be present in an infected leaflet. Their rationale is that the color indicative of CCI can be effectively differentiated from colors associated with other types of infections. Furthermore, when CCI is present, it tends to dominate the leaf, suggesting that there are typically no other pests present. As a result of this insight, the brown pixels represent only the necrotic leaf regions.

K-means has also been proven effective in similar agricultural studies for color segmentation\cite{JAVIDAN2023100081, bashish2011detection}. Therefore, K-means clustering was used to assign each pixel a specific cluster label. Given that our segmentation relied on clear color distinctions—green representing healthy regions and brown indicating infected areas (necrotic regions), Additionally, since we had a prior knowledge of the required clusters, we set the number of clusters (K) to three, corresponding green, brown, and background areas. Using algorithm \ref{alg:Alg2}, we then counted the pixels labeled as green and brown, enabling a clear and accurate quantification of healthy and infected areas. 
Finally, the progression of the disease was calculated by determining the ratio of the necrotic region to the entire leaf region (which includes the total pixel values of both brown and green). Consequently, in addition to the above described the algorithm \ref{alg:Alg2} outlines the process of applying color segmentation, performing K-means clustering, and calculating the progression level.

\subsection{Detect and count Coconut caterpillars}

According to the expert survey during our discussion with the CRISL officers, they currently count the caterpillars manually, by collecting them onto a piece of paper. Based on the number of caterpillars on each coconut leaf, they identify the severity of the infestation. However, despite employing this method, we found it to be not only inaccurate at times but also a tedious task, as the caterpillars are constantly moving., which makes counting them a difficult task. Moreover, this process is time-consuming and prone to human error. Therefore, as one of the objectives of this study, the manual process was automated using image processing techniques. Consequently, the YOLO object detection model \cite{Yolov5,autogyro_yolo_v8_2024,ultralytics2024} was used, as an improvement of the manual process, to count the caterpillars in their natural habitat (leaflet itself) without forcefully extracting them onto paper. Both methodologies were implemented and evaluated to determine which method is more efficient and accurate.

\subsubsection{Counting caterpillars using image processing}

Initially, Gaussian Blur was applied after converting the image into greyscale. Next, the image was converted to a binary image using thresholding after flipping the pixel values using bitwise conversion to see the available objects. All the dark regions were turned black while the light regions were turned white. This was used to prevent a wide range of colors from being present in the image. After that, erosion (i.e., a morphological operation) was used to reduce the features of the image (shrinking the foreground), as it is capable of removing minor noises from images. Erosion can eliminate dust and soil particles that come with the caterpillars. It was also used to separate and identify caterpillars that were closely attached. 

Finally, the connected components were detected, which are defined as neighboring pixel regions with the same input value. Methodically, the algorithm searches for white pixels, assigning each pixel with a unique label ID and applying the same label to all neighboring white pixels iteratively until the entire caterpillar is covered and labeled. Then it moves on to the next accessible white pixel and repeats the process until the entire image is analyzed. Through this process, all the connected components were marked and calculated to detect the number of caterpillars available in the image.

\subsubsection{Counting caterpillars using object detection (YOLO)}

As stated previously, this study used YOLOv5, YOLOv8, and YOLO11 object detection models to calculate caterpillars owing their considerably high accuracy rates   \cite{Mathew2022,9153986,yolov8}. The extra large (x) variants (i.e.,  YOLOv5x, YOLOv8x, and YOLO11x) were employed in this study because the x models achieve significantly higher accuracy compared to other variants, making them well-suited for tasks requiring precise detections \cite{ultralytics2024}. For each version, the \texttt{depth\_multiple} and \texttt{width\_multiple} parameters were maintained at their default values as specified in the original implementations of the respective models \cite{ultralytics2024}. This ensured that the structural integrity of each model remained consistent with its design. The YOLO models were trained on a custom dataset of the coconut caterpillar class. The detection technique was altered to compute the number of caterpillars for each classification.

Similar to Mask R-CNN, the regions of interest are marked based on IoU values. Following non-max suppression, which ensures the object detection algorithm detects each object only once, it returns detected objects with bounding boxes. 

\section{Test setup implementation, results and discussion}

The tests were conducted on a server with the following configurations: GPUs (NVIDIA Tesla T4 \cite{nvidia_tesla_t4}, 16 GB GDDR6 memory @ 300 GBps), CPUs (2 × vCPU), and 24 GB of RAM. The server was implemented in the Google Colab environment using Python version 3.10.12 and performed under the deep learning development framework of TensorFlow and Keras. The training and testing were conducted primarily on the GPU to leverage its parallel processing capabilities for deep learning tasks.

\subsection{Evaluation of WCLWD identification and classification}

\begin{table}[!t]
\centering
\caption{Selecting The Best CNN Architecture by Comparing Testing Accuracies for Each Architecture (\%)}
\label{tab:modelaccuracy}
\renewcommand{\arraystretch}{1.25} 
\begin{tabular}{lcc}
\hline
\textbf{DL Architectures} & \multicolumn{2}{c}{\textbf{WCLWD}}         \\ \cline{2-3} 
             & \textbf{Classification} & \textbf{Symptom Severity} \\ \hline
ResNet50         & 76.92          & 61.32           \\
\textbf{DenseNet121}        & \textbf{90.00} & 88.67           \\
ResNet50V2        & 85.95          & 87.73           \\
\textbf{InceptionResNetV2}     & 81.20          & \textbf{97.00}         \\
InceptionV3        & 84.62          & 77.35           \\
MobileNetV2        & 78.63          & 95.28           \\
Xception         & 84.62          & 82.07           \\
VGG16           & -            & 90.56           \\ \hline
\end{tabular}
\end{table}

As shown in Table \ref{tab:SummaryofDataSamples}, the dataset was initially divided into training and validation. Then a separate dataset was collected for the testing purposes. The models for WCLWD, and its symptom severity were trained using several architectures as summarized in Table \ref{tab:modelaccuracy}, and by considering the accuracy, the best architecture was chosen. The loss values were considered to define the model's performance after each iteration of optimization. Table \ref{tab:modelaccuracy} shows the testing accuracies for WCLWD classification and symptom severity.

Accuracy \eqref{eq:accuracy}, confusion matrix, precision \eqref{eq:precision}, recall \eqref{eq:recall}, and F1-score \eqref{eq:F1_score}, as explained in \cite{Alzubaidi2021}, are used to analyze the performance of both WCLWD classification and symptom severity models. The Accuracy is defined as the ratio of correct predictions to the total number of predictions, as given by \eqref{eq:accuracy}, where True Positive (TP), False Positive (FP), True Negative (TN), and False Negative (FN) represent cases when the model predicts the positive class as positive (i.e., TP) or as negative (i.e., FN), and predicts the negative class as positive (i.e., FP) or as negative (i.e., TN), respectively.

\begin{equation}
\label{eq:accuracy}
\text{Accuracy} = \frac{TP + TN}{TP + TN + FP + FN}
\end{equation}

The Precision \eqref{eq:precision} is defined as the ratio of correctly predicted positive observations to the total predicted positive observations:

\begin{equation}
\label{eq:precision}
Precision = \frac{TP}{TP + FP}
\end{equation}

The Recall \eqref{eq:recall}, also known as sensitivity, is the ratio of correctly predicted positive observations to all actual positives:

\begin{equation}
\label{eq:recall}
Recall = \frac{TP}{TP + FN}
\end{equation}

The F1 score \eqref{eq:F1_score} is the harmonic mean of Precision and Recall, providing a balance between the two metrics:

\begin{equation}
\label{eq:F1_score}
F_{1\text{ score}} = 2 \times \frac{Precision \times Recall}{Precision + Recall}
\end{equation}

\begin{figure}[!t]
  \centering
  \begin{subfigure}[b]{0.45\linewidth}
    \centering
    \includegraphics[width=\linewidth]{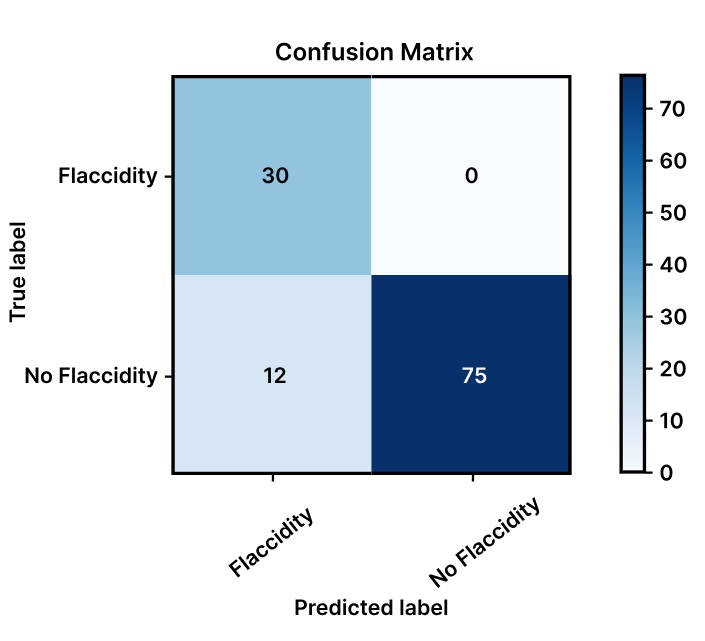}
    \caption{Classification}
    \label{fig:ConfusionMatrixClassification}
  \end{subfigure}
  \hfill
  \begin{subfigure}[b]{0.45\linewidth}
    \centering
    \includegraphics[width=\linewidth]{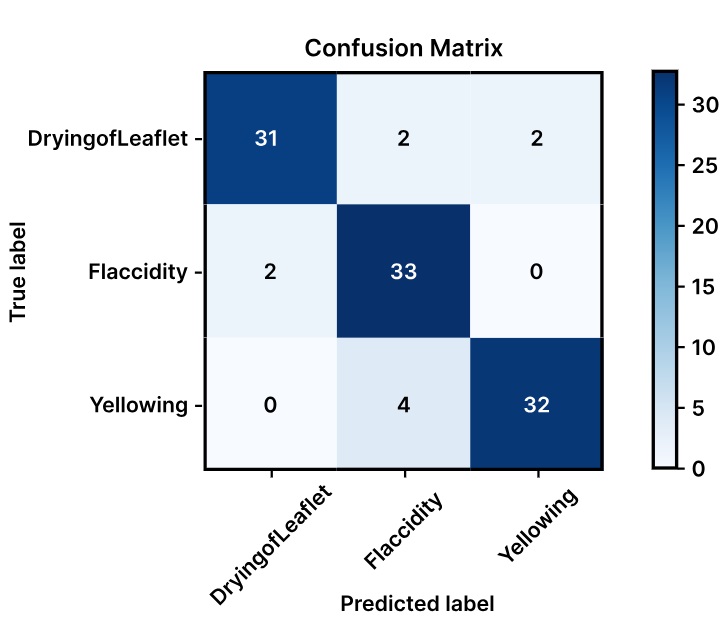}
    \caption{Symptom severity}
    \label{fig:ConfusionMatrixSymptomSeverity}
  \end{subfigure}
  \caption{Confusion matrices WCLWD}
  \label{fig:ConfusionMatrix}
\end{figure}

\begin{figure*}[!t]
  \centering
  \begin{subfigure}[b]{0.45\linewidth}
    \centering
    \includegraphics[width=\linewidth]{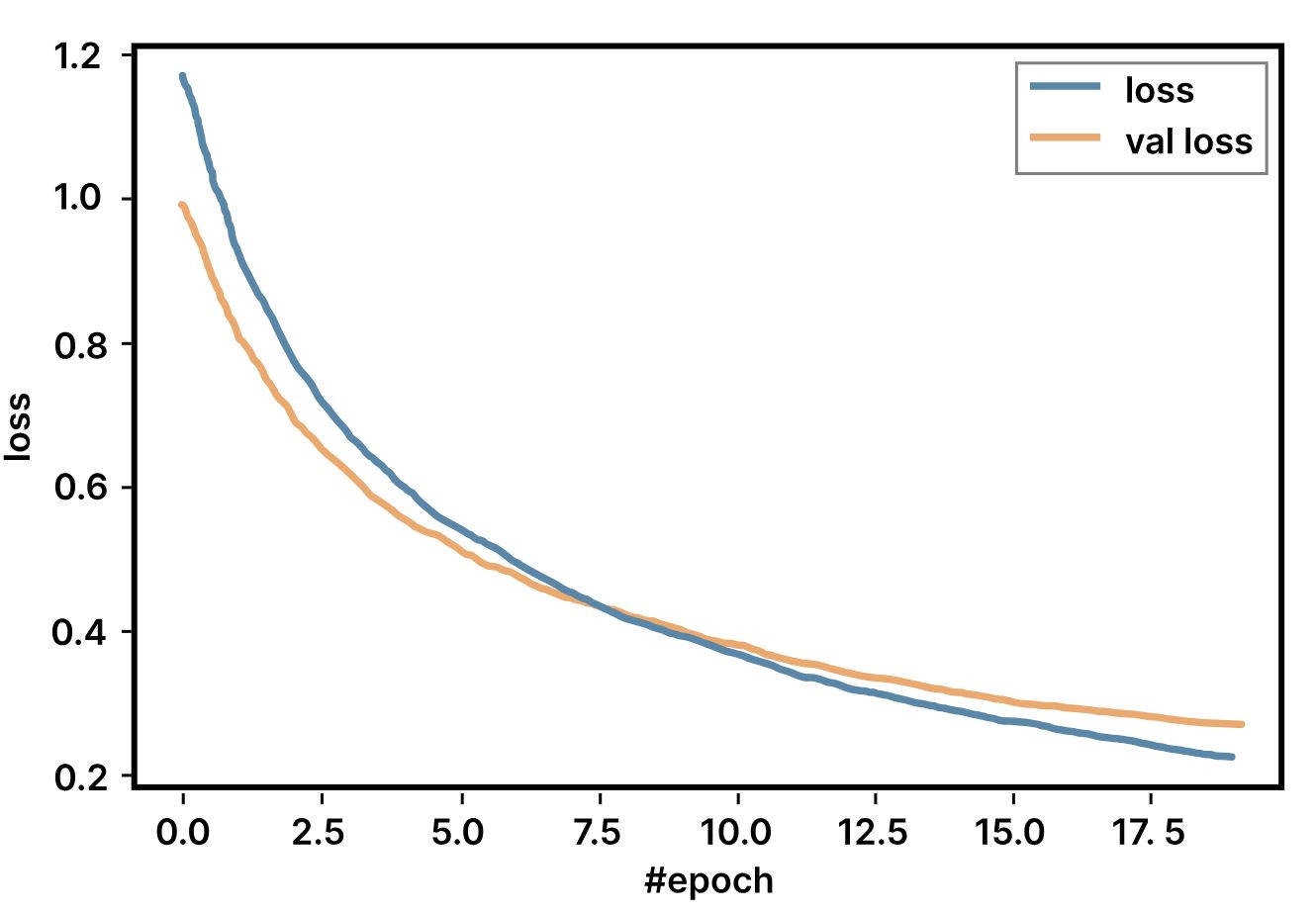}
    \caption{}
    \label{fig:lossWCLWD}
  \end{subfigure}
  \hfill
  \begin{subfigure}[b]{0.45\linewidth}
    \centering
    \includegraphics[width=\linewidth]{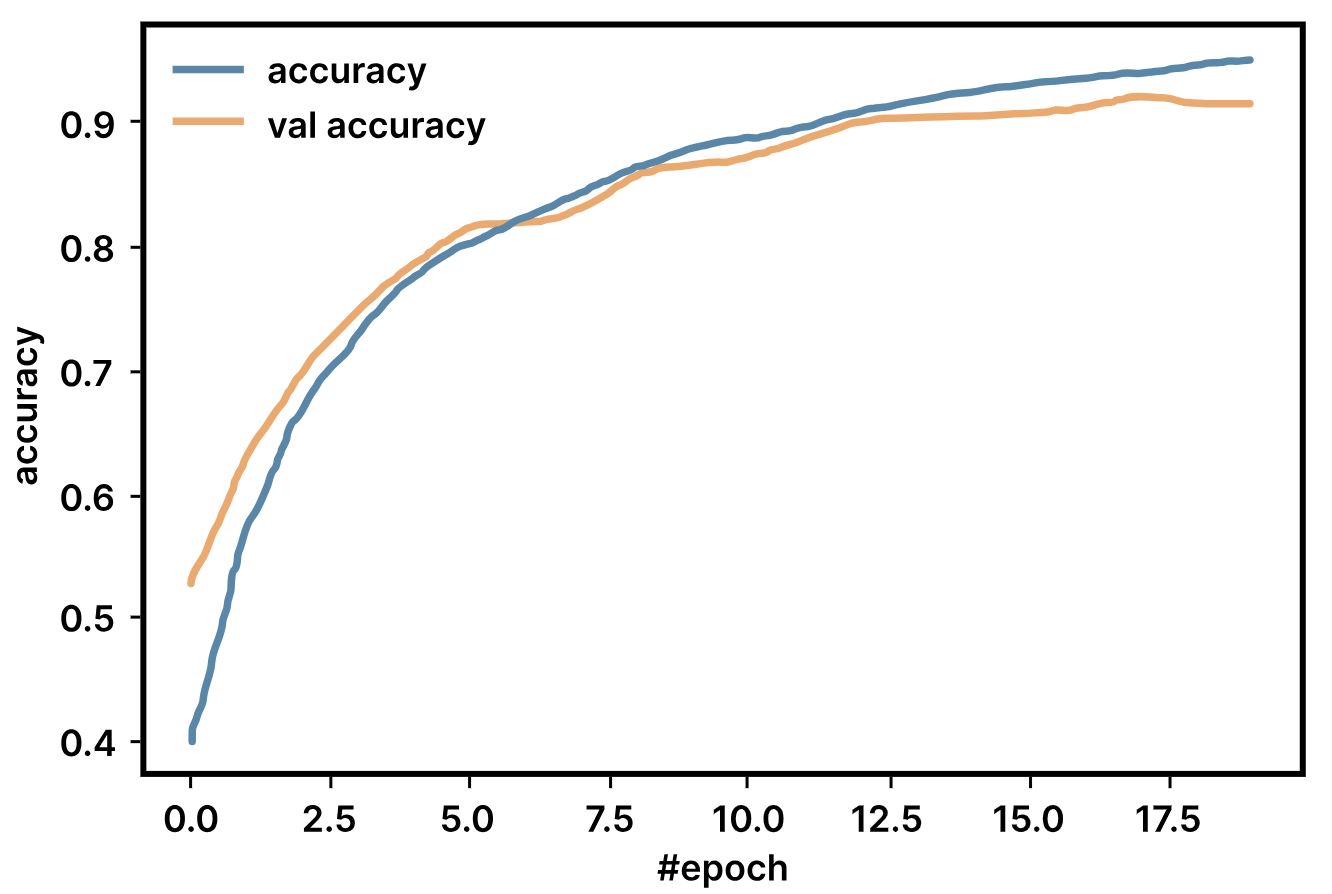}
    \caption{}
    \label{fig:accuracyWCLWD}
  \end{subfigure}
  \vfill
  \begin{subfigure}[b]{0.45\linewidth}
    \centering
    \includegraphics[width=\linewidth]{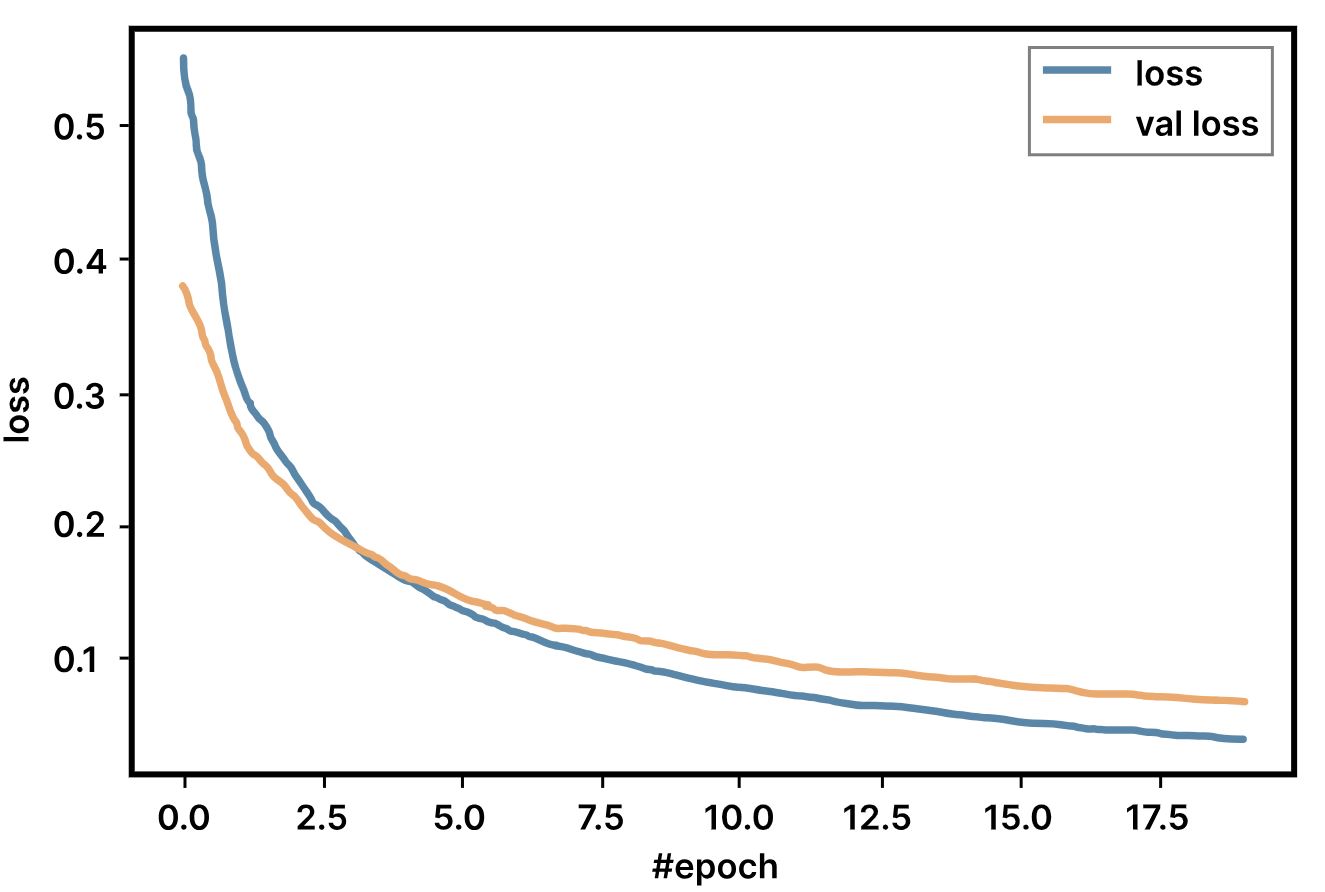}
    \caption{}
    \label{fig:lossSymptomSeverity}
  \end{subfigure}
  \hfill
  \begin{subfigure}[b]{0.45\linewidth}
    \centering
    \includegraphics[width=\linewidth]{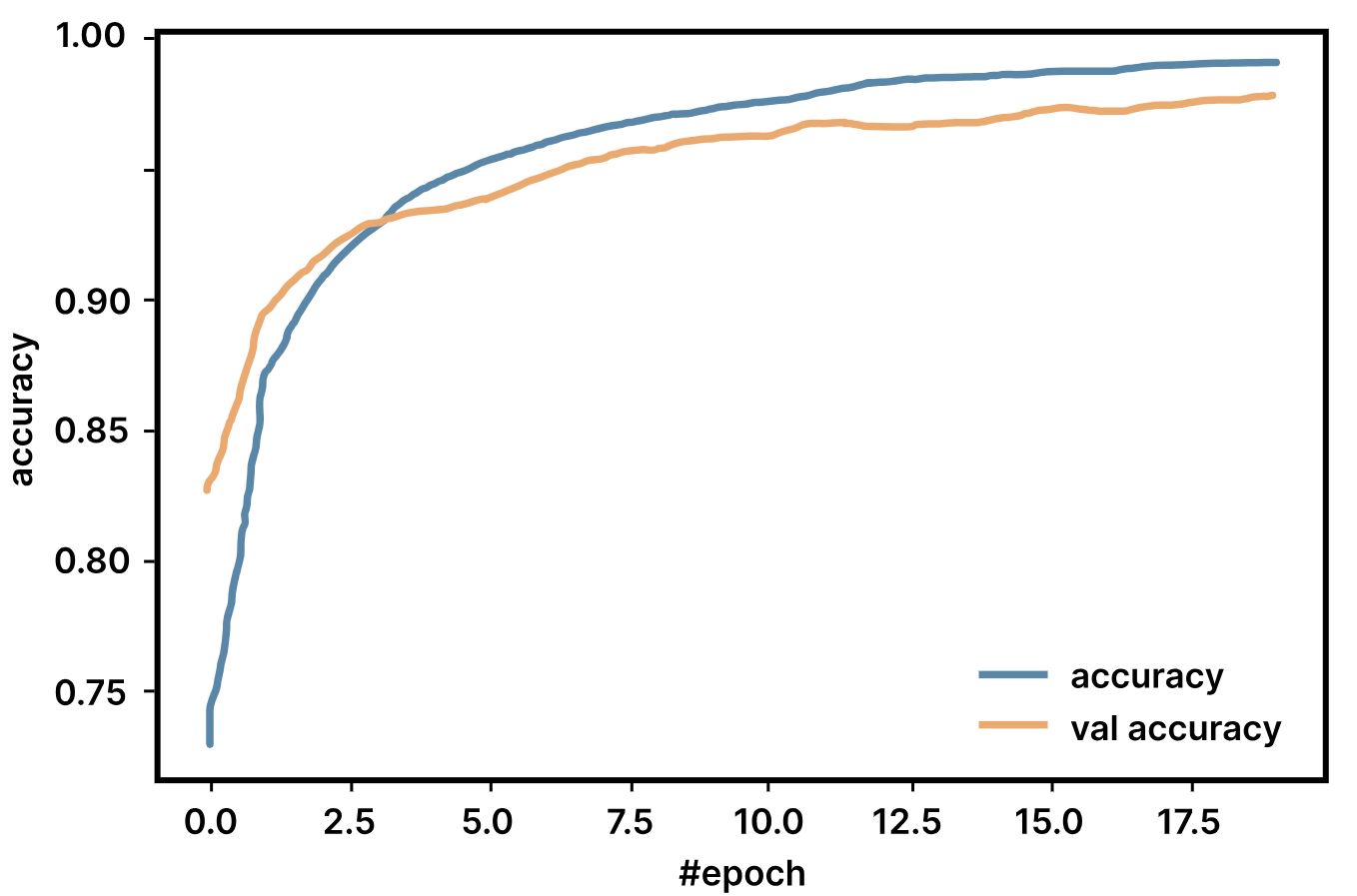}
    \caption{}
    \label{fig:accuracySymptomSeverity}
  \end{subfigure}

  \caption{{Loss and accuracy plots: a. Loss curve for WCLWD classification, b. Accuracy curve for WCLWD classification, c. Loss curve for symptom severity, d. Accuracy curve for symptom severity}}
  \label{fig:lossaccuracy}
\end{figure*}

The WCLWD classification model using the DenseNet121 achieved 99.03\% training accuracy and 90\% testing accuracy which performs the best test accuracy among several architectural models. Moreover, the InceptionResNetV2-based symptom severity model achieved training and testing accuracies of 99.68\% and 97\%, respectively. Figure \ref{fig:ConfusionMatrix} illustrates the confusion matrices related to WCLWD classification (a) and symptom severity (b), which help to evaluate the performance of the trained models. In addition, Fig. \ref{fig:lossaccuracy} represents the loss and accuracy plots of both models DenseNet121 (a and b) and InceptionResNetV2 (c and d) respectively.

The precision, recall, and F1-score of the DenseNet121 model used for WCLWD identification and classification were 71\%, 100\%, and 83\%, respectively. Similarly, the InceptionResNetV2 model, which was used to determine symptom severity achieved 99\%, 99\%, and 98\%.

\begin{figure}[!t]
  \centering
  \begin{subfigure}[b]{0.24\linewidth}
    \centering
    \includegraphics[width=\linewidth]{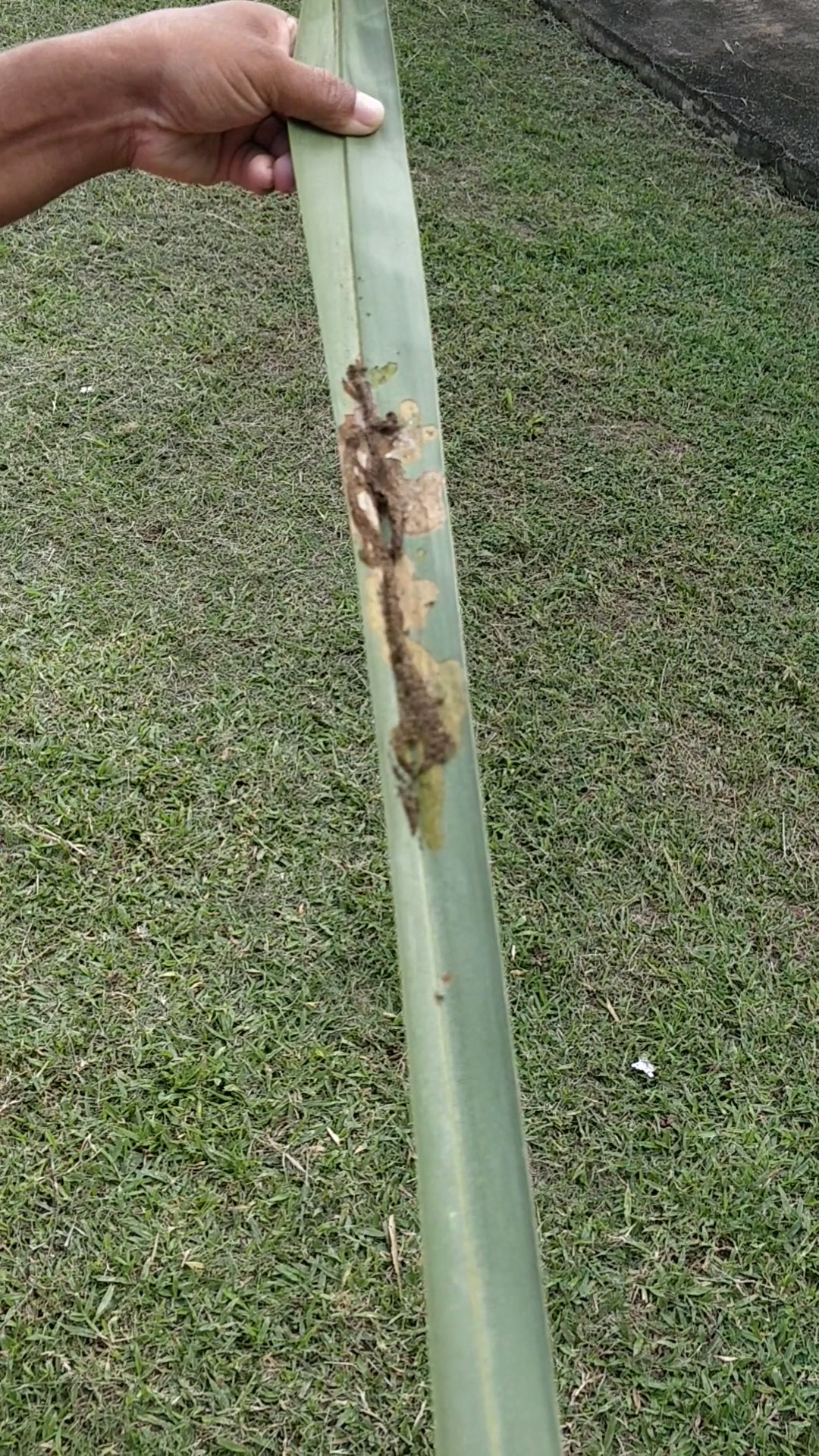}
    \caption{}
    \label{fig:original_image}
  \end{subfigure}\hfill
  \begin{subfigure}[b]{0.24\linewidth}
    \centering
    \includegraphics[width=\linewidth]{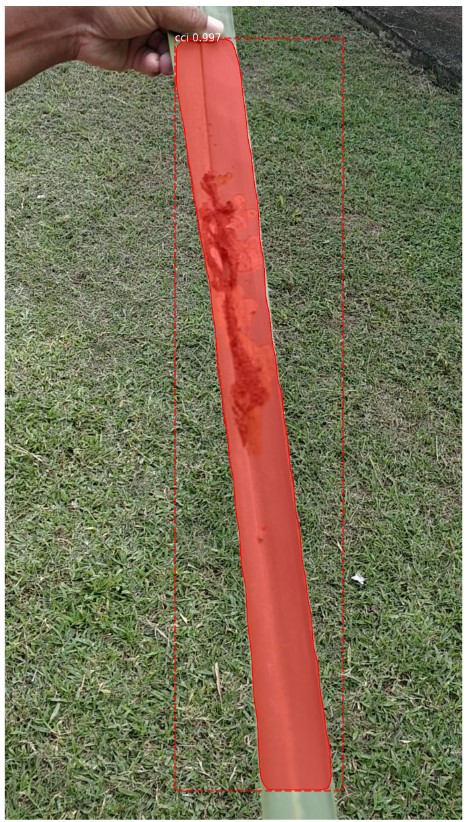}
    \caption{}
    \label{fig:masking}
  \end{subfigure}\hfill
  \begin{subfigure}[b]{0.24\linewidth}
    \centering
    \includegraphics[width=\linewidth]{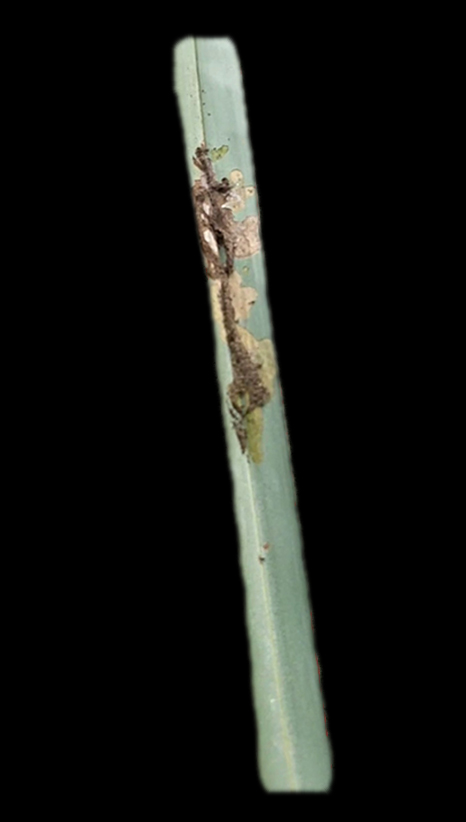}
    \caption{}
    \label{fig:crop_segmentation}
  \end{subfigure}\hfill
  \begin{subfigure}[b]{0.24\linewidth}
    \centering
    \includegraphics[width=\linewidth]{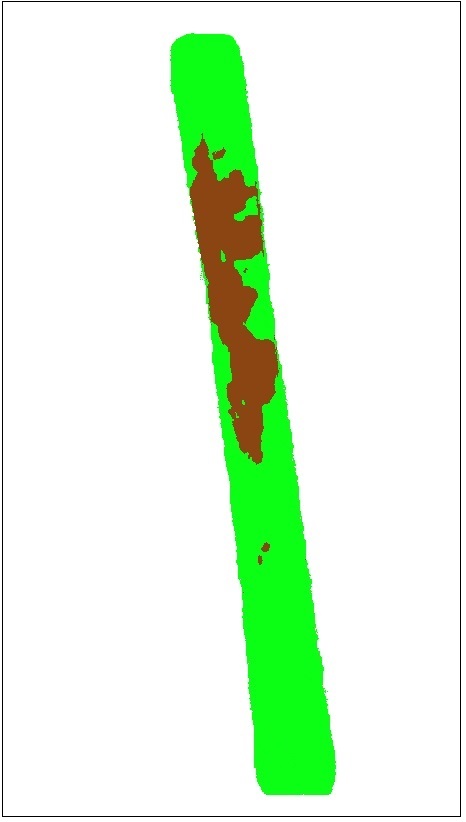}
    \caption{}
    \label{fig:color_segmentation}
  \end{subfigure}
  \caption{Process of calculating the extent of damage: a. Original image, b. Masking, c.
Crop segmentation, d. Color segmentation}
  \label{fig:CCIprocess}
\end{figure}

\begin{figure}[!ht]
  \centering
  \begin{subfigure}[b]{0.8\linewidth}
    \centering
    \includegraphics[width=\linewidth]{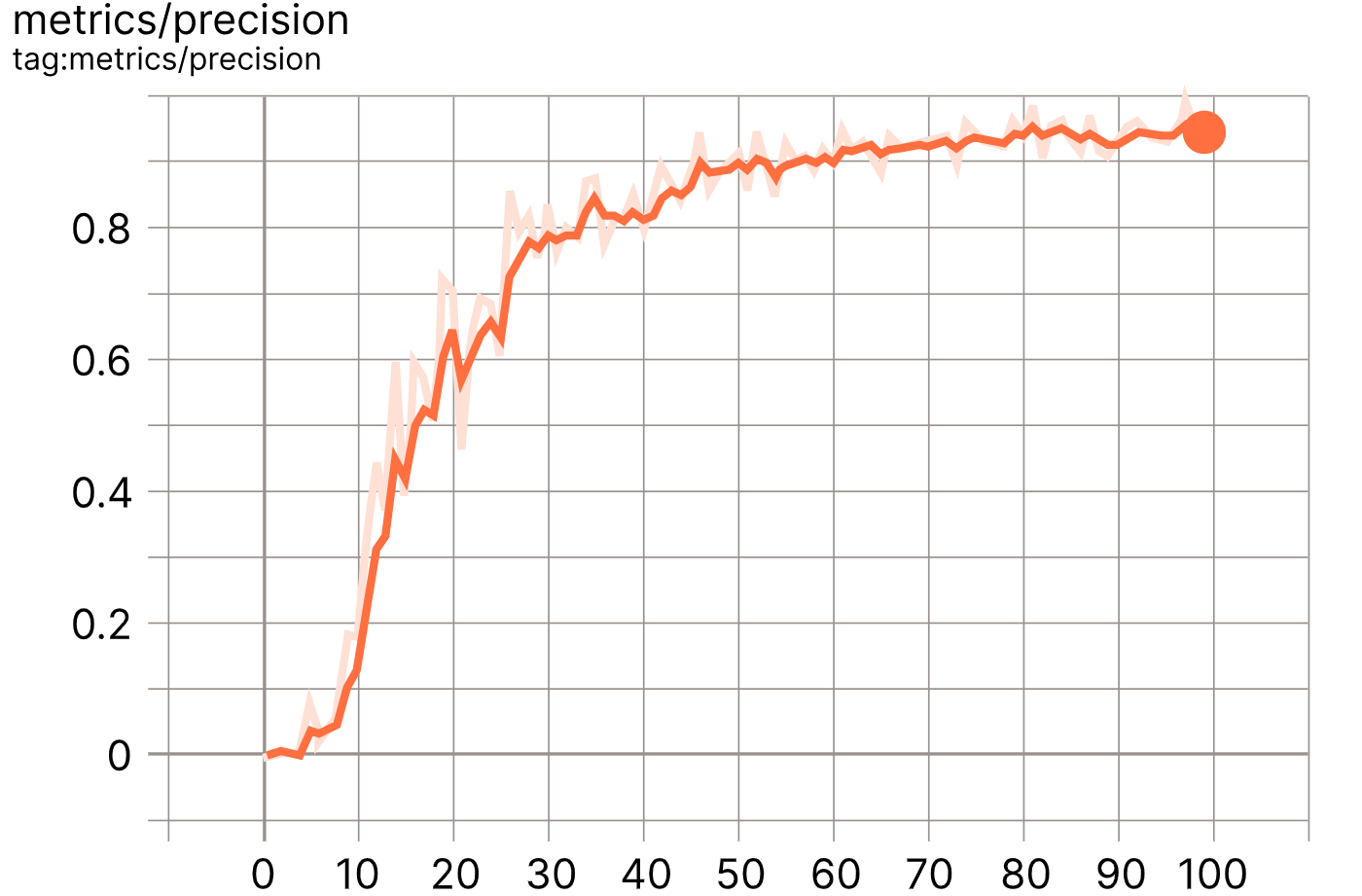}
    \caption{Precision}
    \label{fig:MaskRCNNPrecision}
  \end{subfigure}
  \vfill
  \begin{subfigure}[b]{0.8\linewidth}
    \centering
    \includegraphics[width=\linewidth]{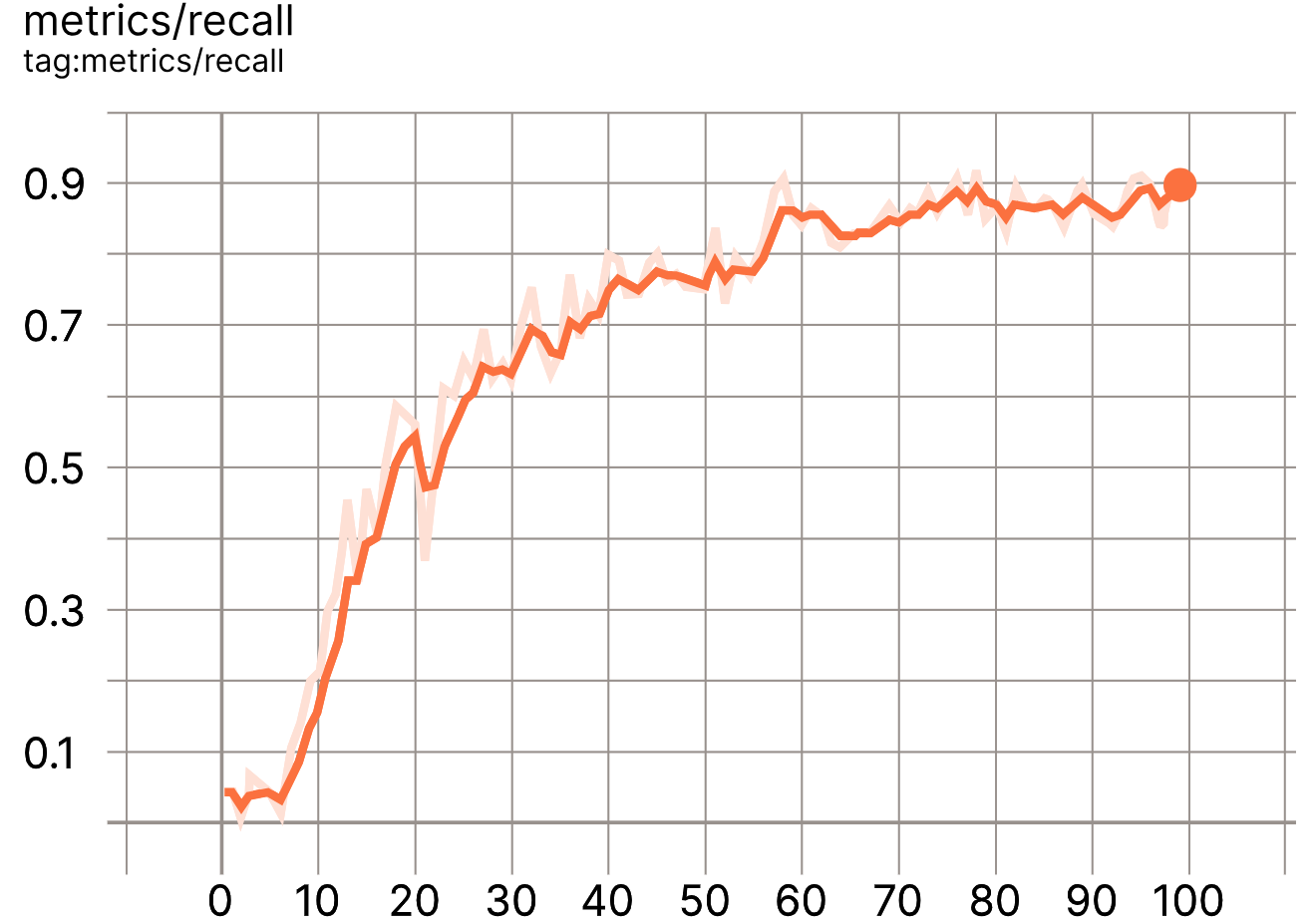}
    \caption{Recall}
    \label{fig:MaskRCNNRecall}
  \end{subfigure}
  \vfill
  \begin{subfigure}[b]{0.8\linewidth}
    \centering
    \includegraphics[width=\linewidth]{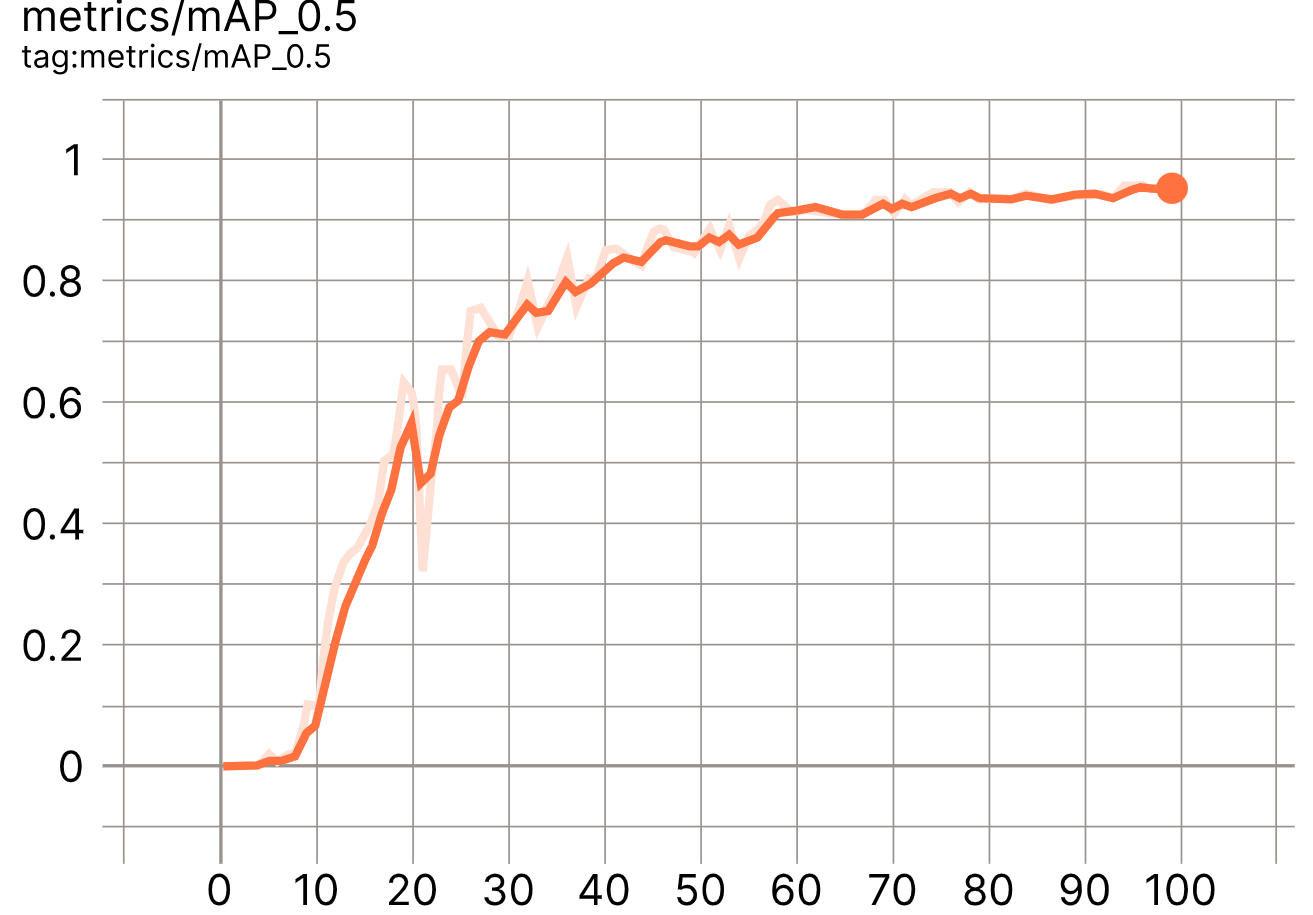}
    \caption{mAP}
    \label{fig:MaskRCNNmAP}
  \end{subfigure}
  \caption{Accuracy graphs for Mask R-CNN model: a. Precision graph, b. Recall graph, c. mAP graph.}
  \label{fig:MaskRCNNGraph}
\end{figure}

\begin{figure*}[!t]
  \centering
  \begin{subfigure}[b]{0.22\linewidth}
    \centering
    \includegraphics[width=\linewidth]{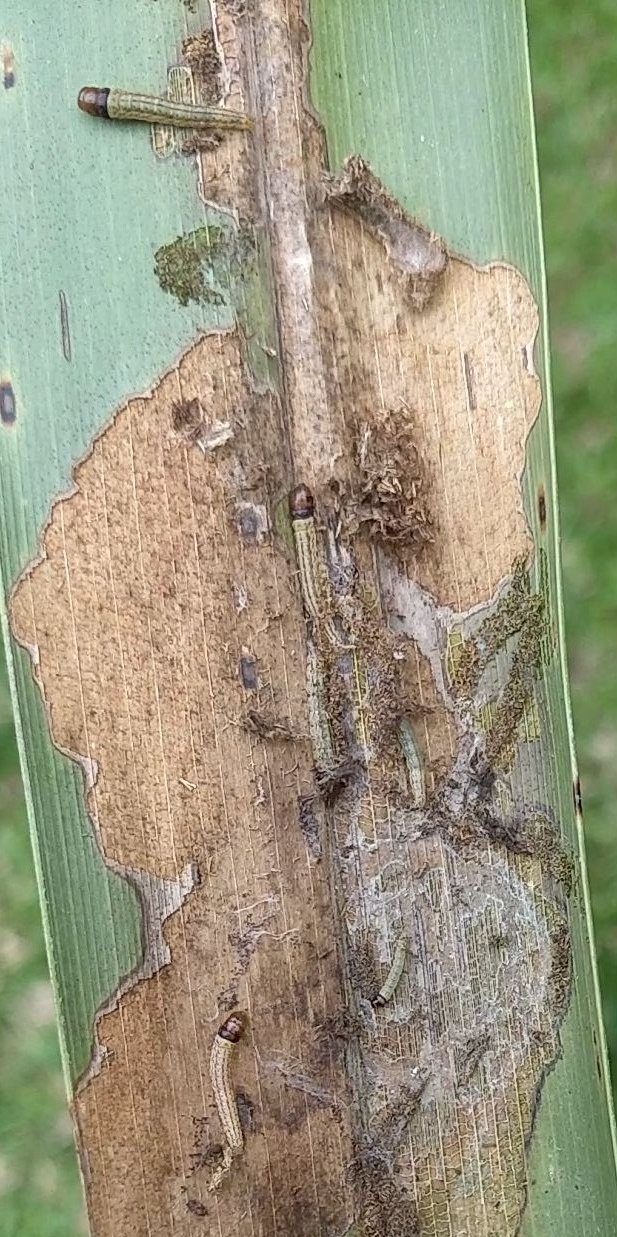}
    \caption{}
    \label{fig:OriginalImageCCI}
  \end{subfigure}
  \hfill
  \begin{subfigure}[b]{0.22\linewidth}
    \centering
    \includegraphics[width=\linewidth]{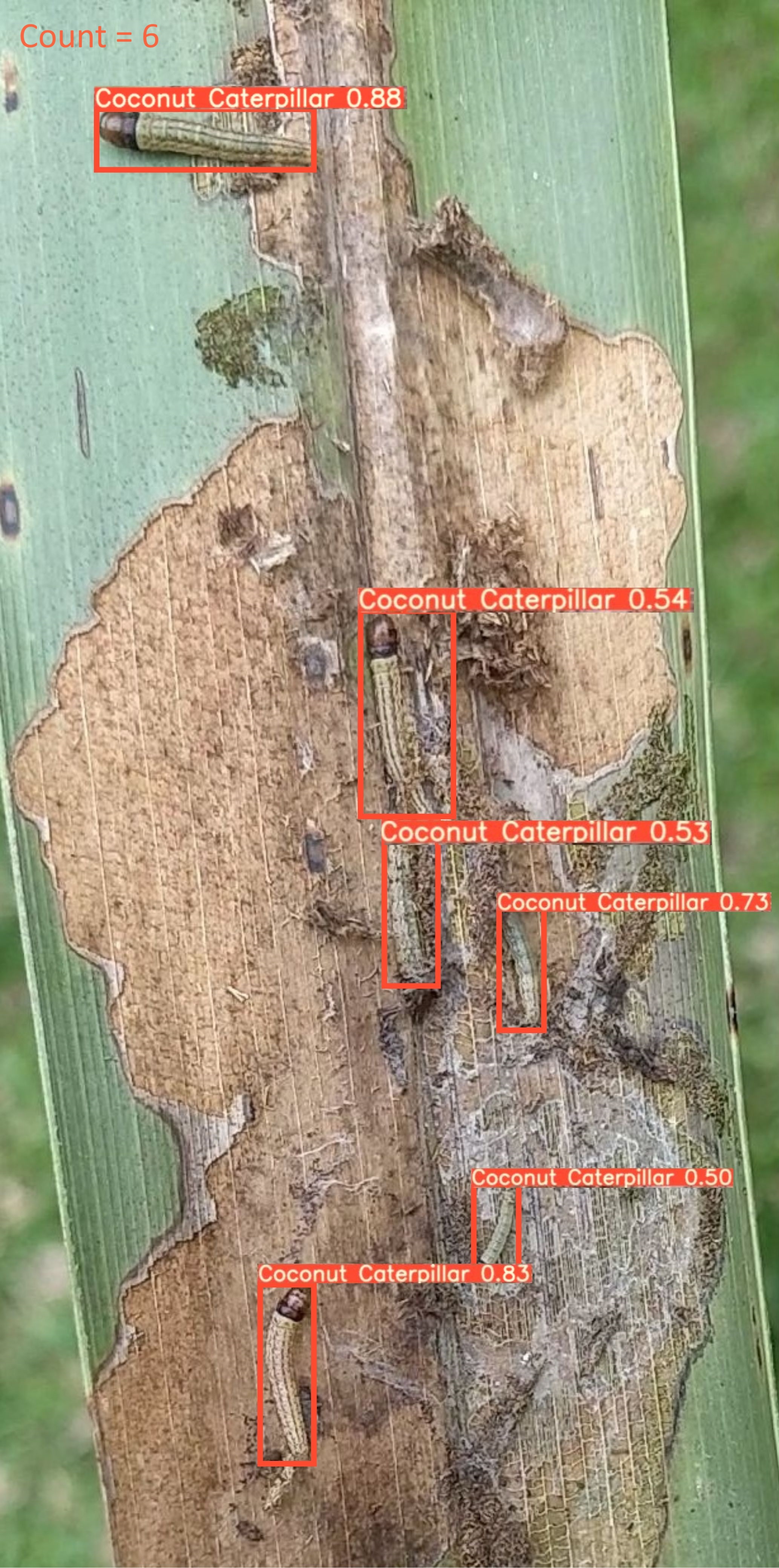}
    \caption{}
    \label{fig:DetectedImageCCI}
  \end{subfigure}
  \hfill
  \begin{subfigure}[b]{0.22\linewidth}
    \centering
    \includegraphics[width=\linewidth]{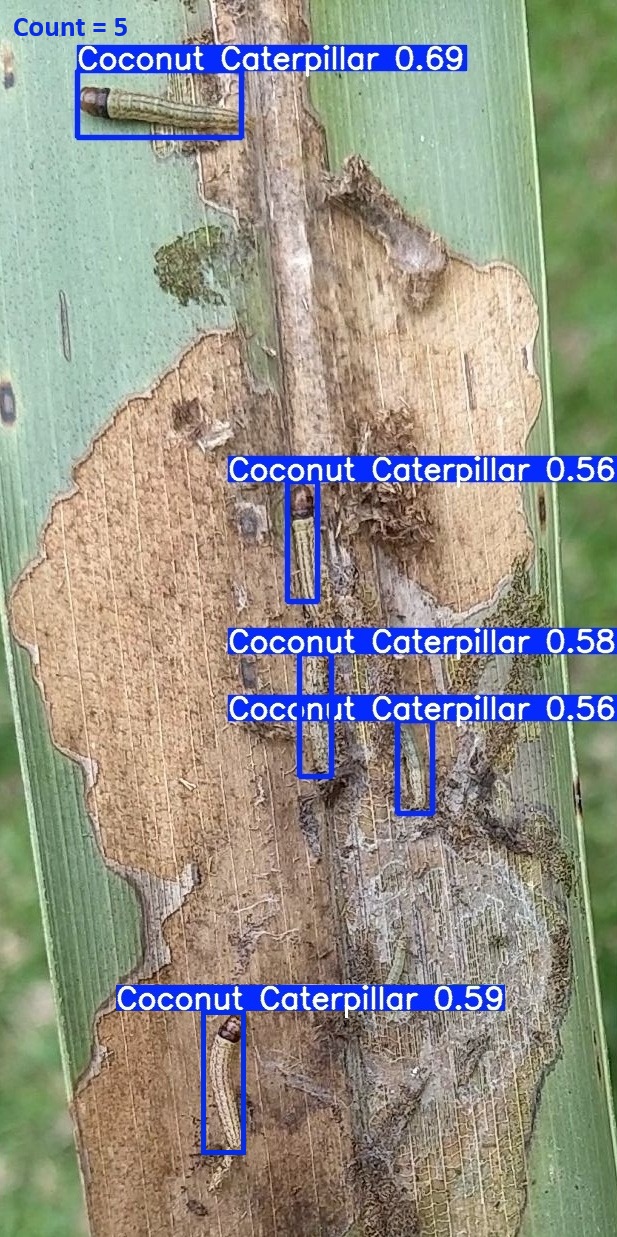}
    \caption{}
    \label{fig:SegmentedImageCCI}
  \end{subfigure}
  \hfill
  \begin{subfigure}[b]{0.22\linewidth}
    \centering
    \includegraphics[width=\linewidth]{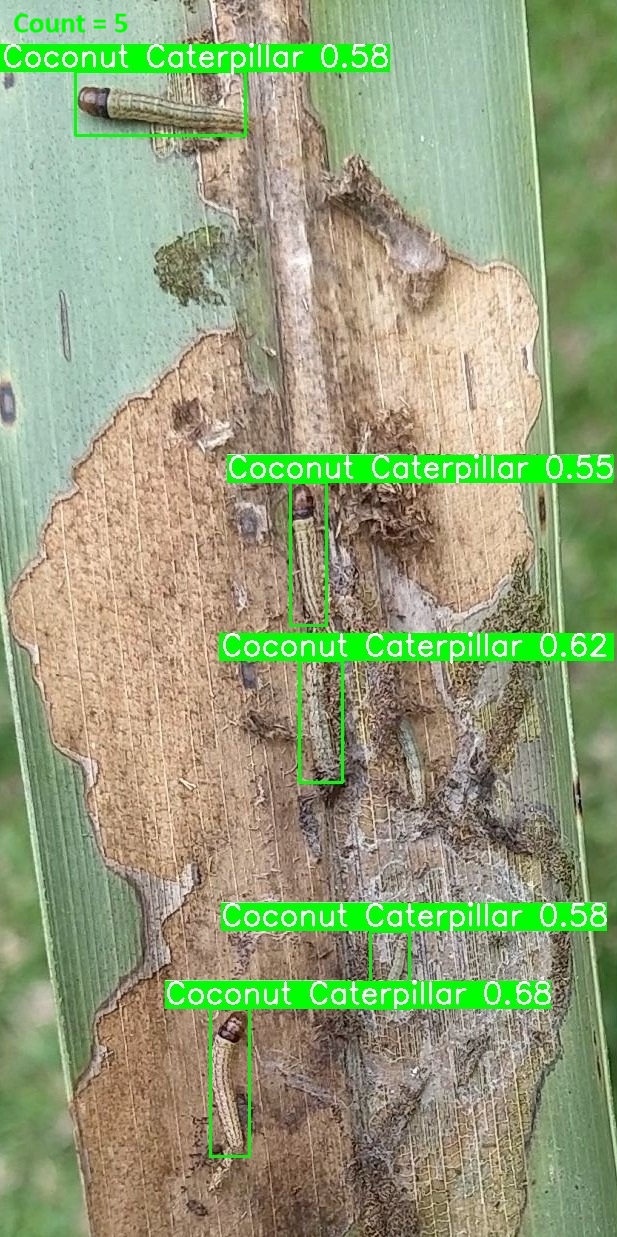}
    \caption{}
    \label{fig:MaskedImageCCI}
  \end{subfigure}
  \caption{Counting the number of caterpillars using YOLO models: 
  (a) Original image with 6 caterpillars, 
  (b) YOLOv5x detection (6/6 correctly identified), 
  (c) YOLOv8x detection (5/6 correctly identified), 
  (d) YOLOv11x detection (5/6 correctly identified.}
  \label{fig:catcountImg}
\end{figure*}

\begin{figure}[!t]
  \centering
  \begin{subfigure}[b]{0.32\linewidth}
    \centering
    \includegraphics[width=\linewidth]{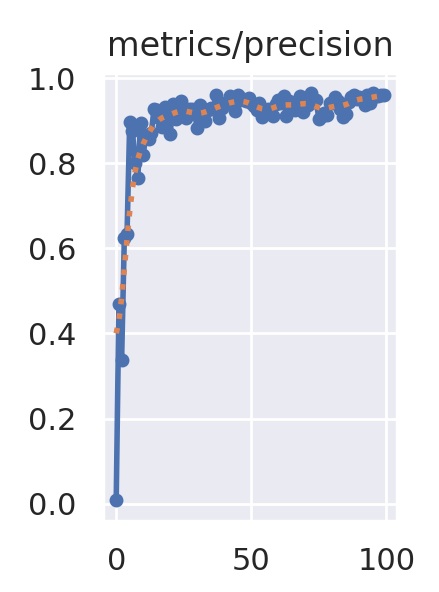}
    \caption{Precision }
    \label{fig:YoloPrecision}
  \end{subfigure}\hfill
  \begin{subfigure}[b]{0.32\linewidth}
    \centering
    \includegraphics[width=\linewidth]{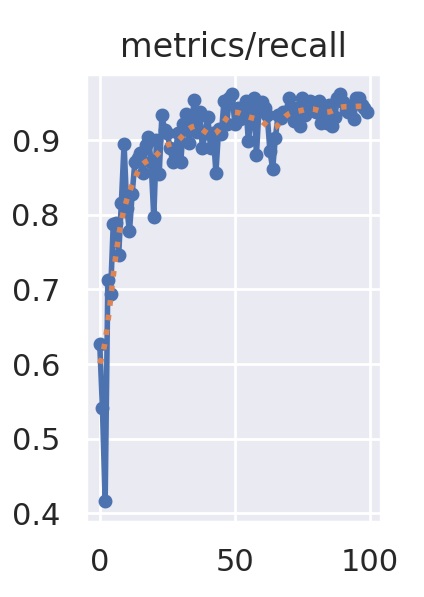}
    \caption{Recall }
    \label{fig:YoloRecall}
  \end{subfigure}\hfill
  \begin{subfigure}[b]{0.32\linewidth}
    \centering
    \includegraphics[width=\linewidth]{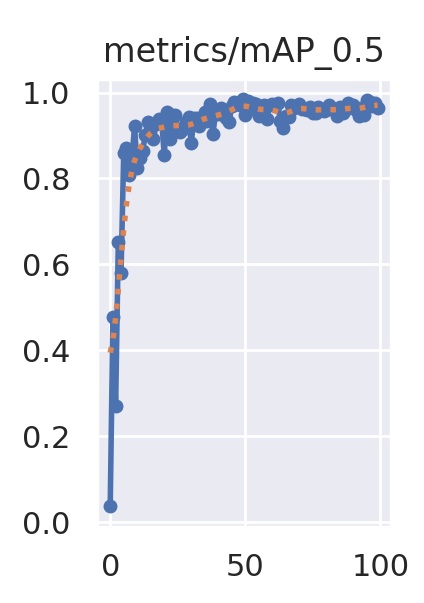}
    \caption{mAP }
    \label{fig:YolomAP}
  \end{subfigure}
  \caption{Graphs for YOLOv5 model: a. Precision graph, b. Recall graph, c. mAP graph}
  \label{fig:YoloGraph5}
\end{figure}

\begin{figure}[!t]
  \centering
  \begin{subfigure}[b]{0.32\linewidth}
    \centering
    \includegraphics[width=\linewidth]{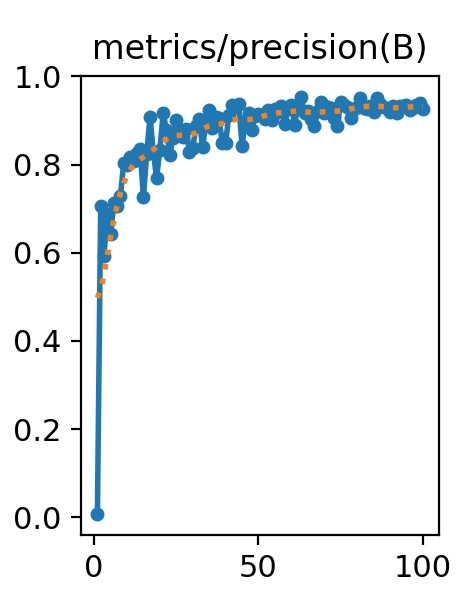}
    \caption{Precision }
    \label{fig:YoloPrecision}
  \end{subfigure}\hfill
  \begin{subfigure}[b]{0.32\linewidth}
    \centering
    \includegraphics[width=\linewidth]{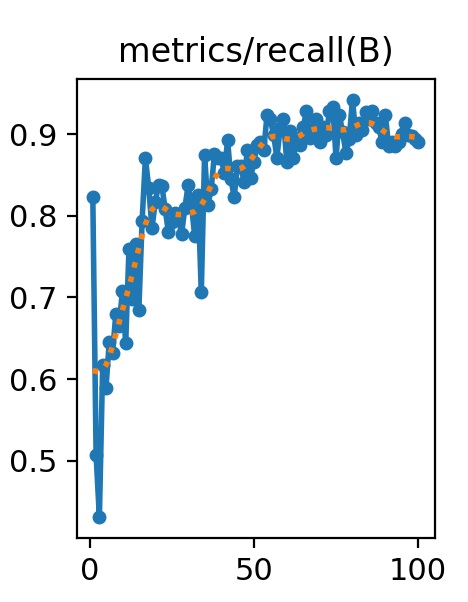}
    \caption{Recall }
    \label{fig:YoloRecall}
  \end{subfigure}\hfill
  \begin{subfigure}[b]{0.32\linewidth}
    \centering
    \includegraphics[width=\linewidth]{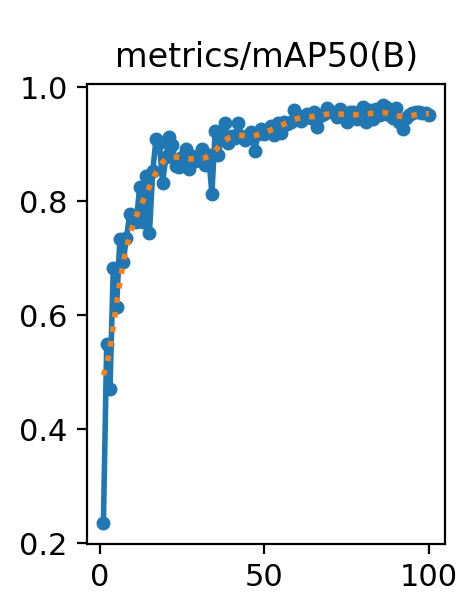}
    \caption{mAP }
    \label{fig:YolomAP}
  \end{subfigure}
  \caption{Graphs for YOLOv8 model: a. Precision graph, b. Recall graph, c. mAP graph}
  \label{fig:YoloGraph8}
\end{figure}

\begin{figure}[!t]
  \centering
  \begin{subfigure}[b]{0.32\linewidth}
    \centering
    \includegraphics[width=\linewidth]{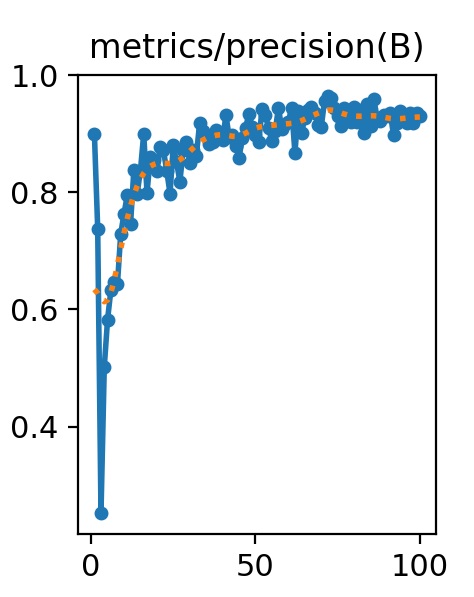}
    \caption{Precision }
    \label{fig:YoloPrecision}
  \end{subfigure}\hfill
  \begin{subfigure}[b]{0.32\linewidth}
    \centering
    \includegraphics[width=\linewidth]{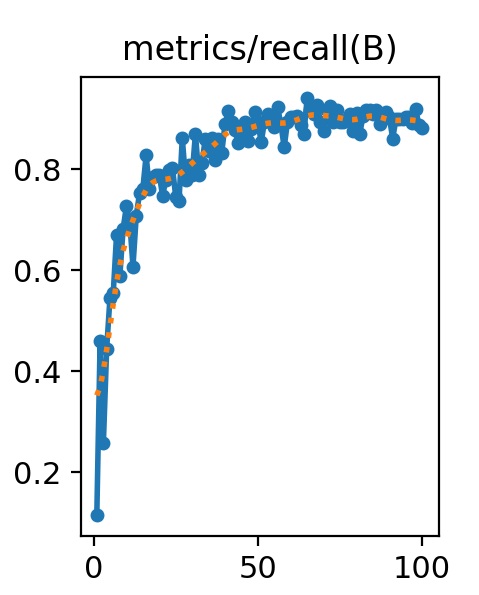}
    \caption{Recall }
    \label{fig:YoloRecall}
  \end{subfigure}\hfill
  \begin{subfigure}[b]{0.32\linewidth}
    \centering
    \includegraphics[width=\linewidth]{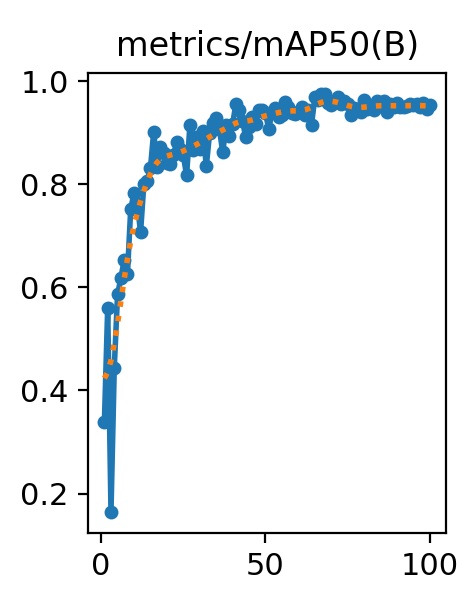}
    \caption{mAP }
    \label{fig:YolomAP}
  \end{subfigure}
  \caption{Graphs for YOLO11 model: a. Precision graph, b. Recall graph, c. mAP graph}
  \label{fig:YoloGraph11}
\end{figure}

\begin{figure}[!t]
  \centering
  \begin{subfigure}[b]{0.24\linewidth}
    \centering
    \includegraphics[width=\linewidth]{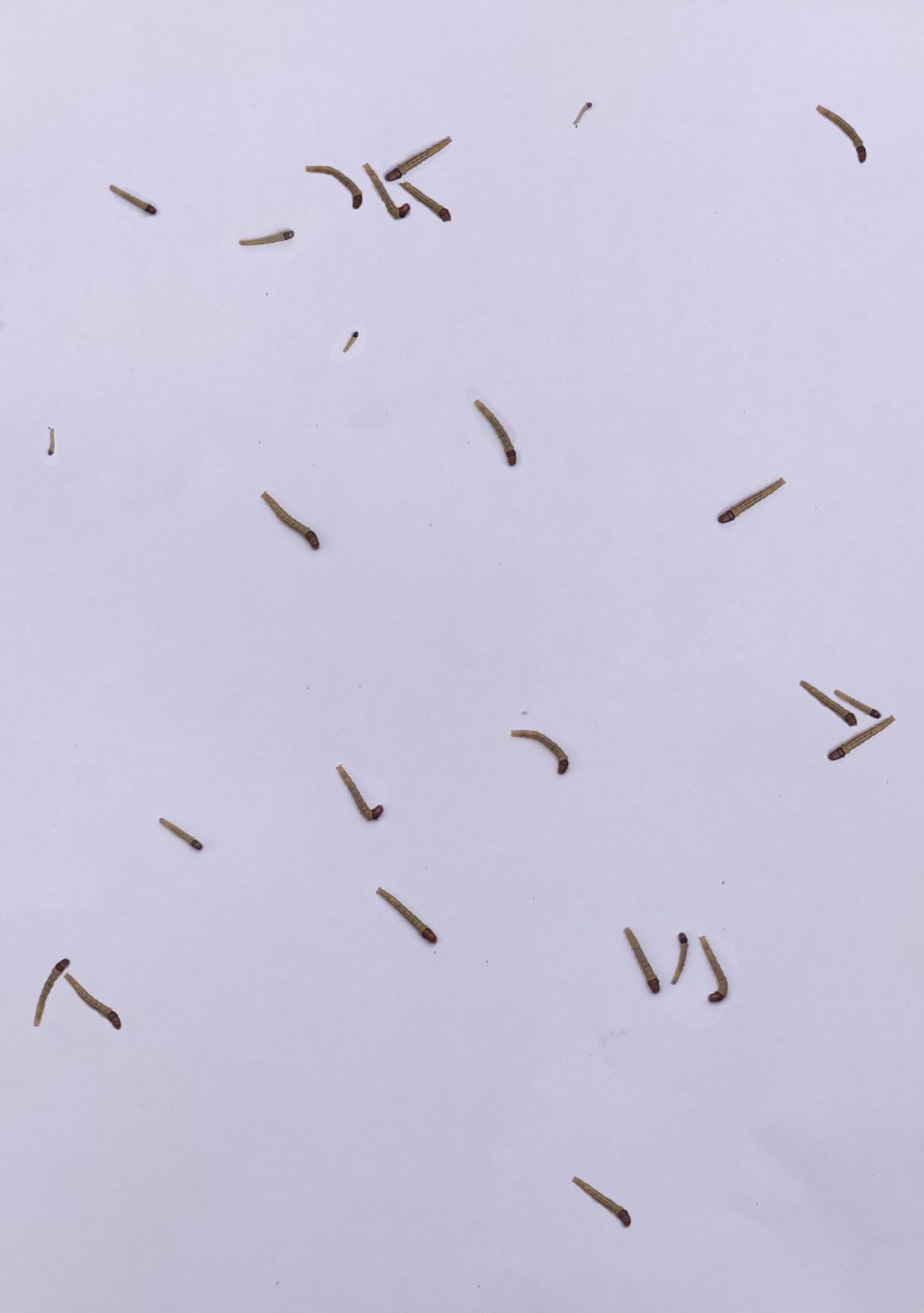}
    \caption{}
    \label{fig:catcount_original}
  \end{subfigure}\hfill
  \begin{subfigure}[b]{0.24\linewidth}
    \centering
    \includegraphics[width=\linewidth]{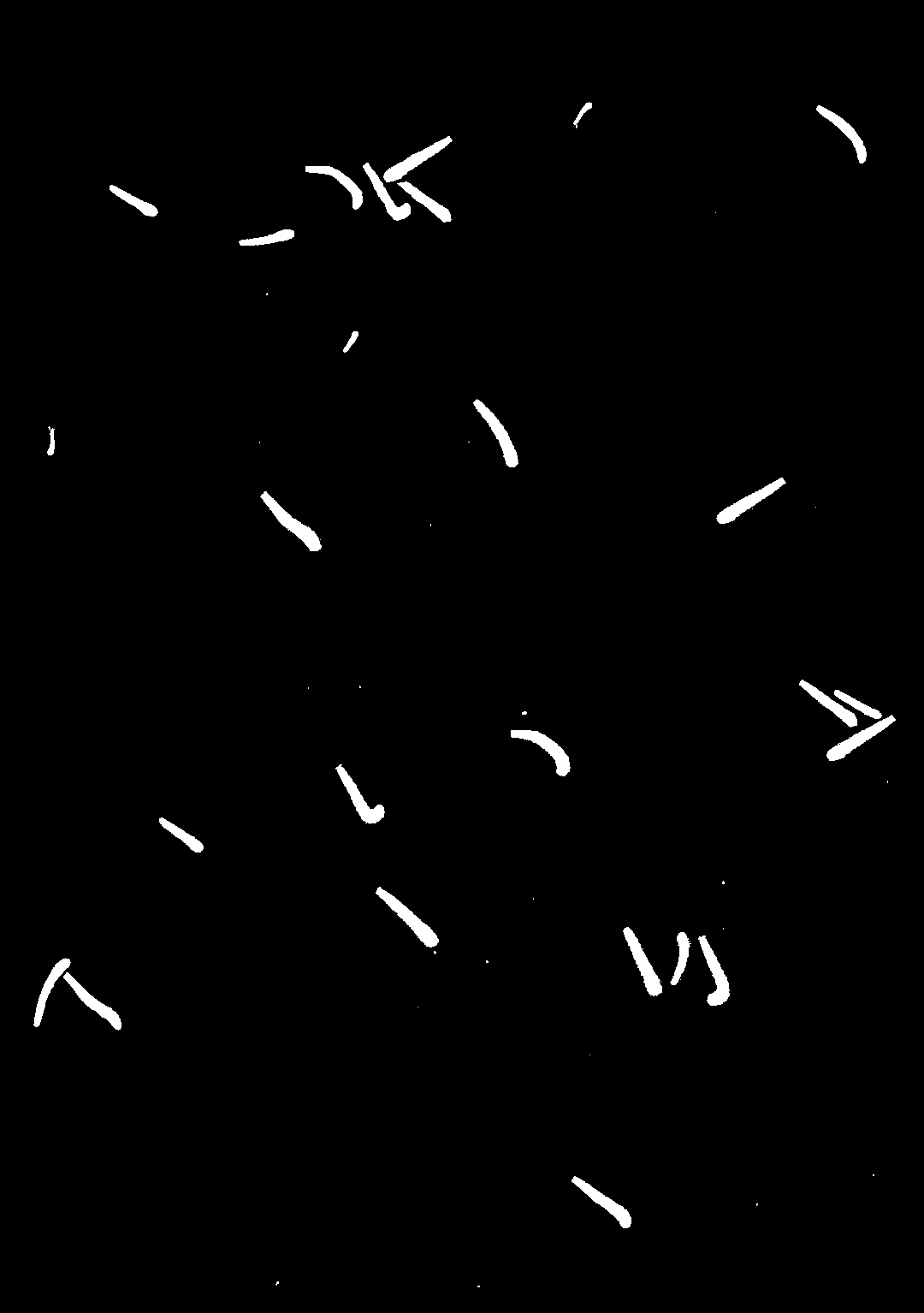}
    \caption{}
    \label{fig:catcount_thresholding}
  \end{subfigure}\hfill
  \begin{subfigure}[b]{0.24\linewidth}
    \centering
    \includegraphics[width=\linewidth]{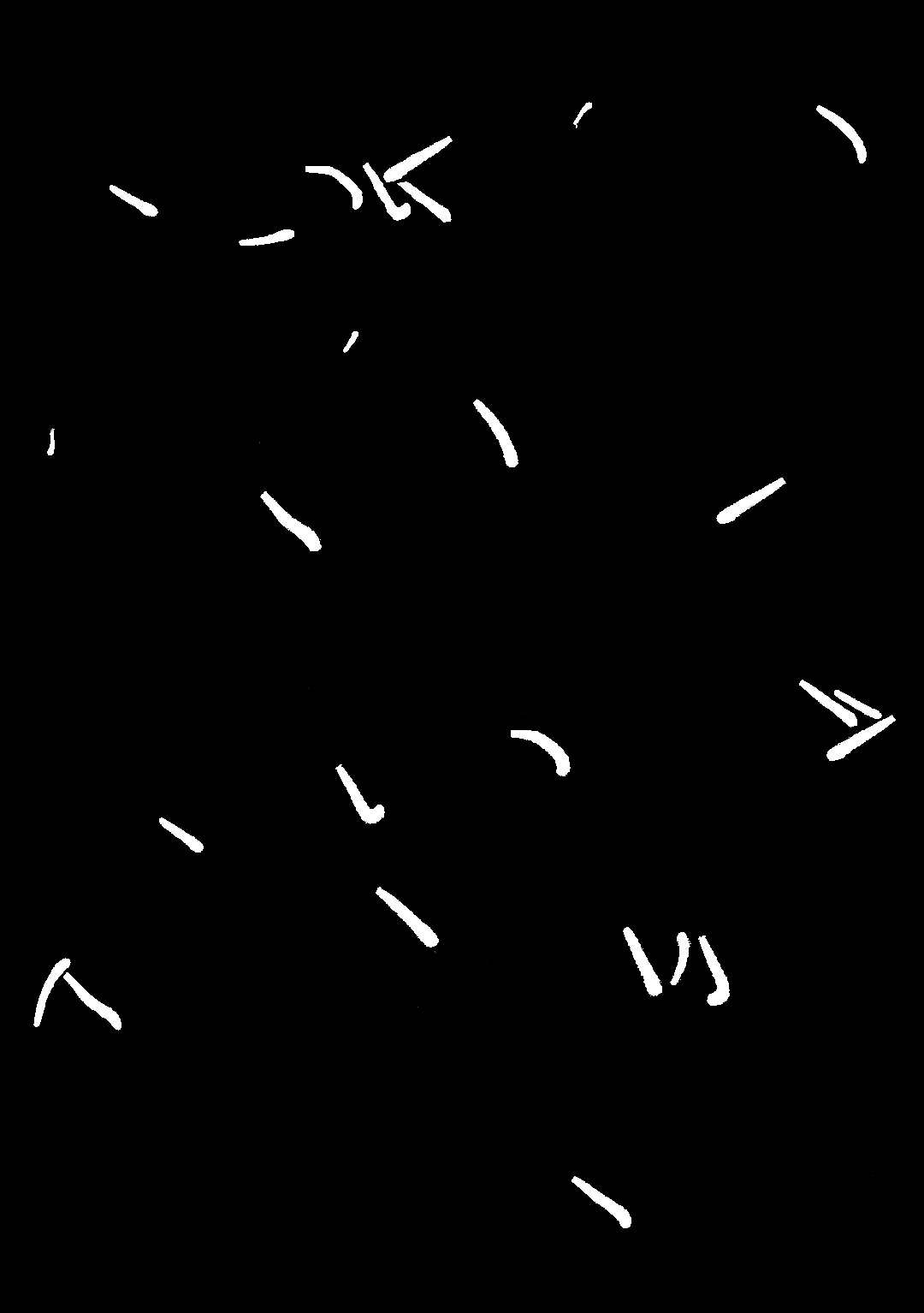}
    \caption{}
    \label{fig:catcount_erosion}
  \end{subfigure}\hfill
  \begin{subfigure}[b]{0.24\linewidth}
    \centering
    \includegraphics[width=\linewidth]{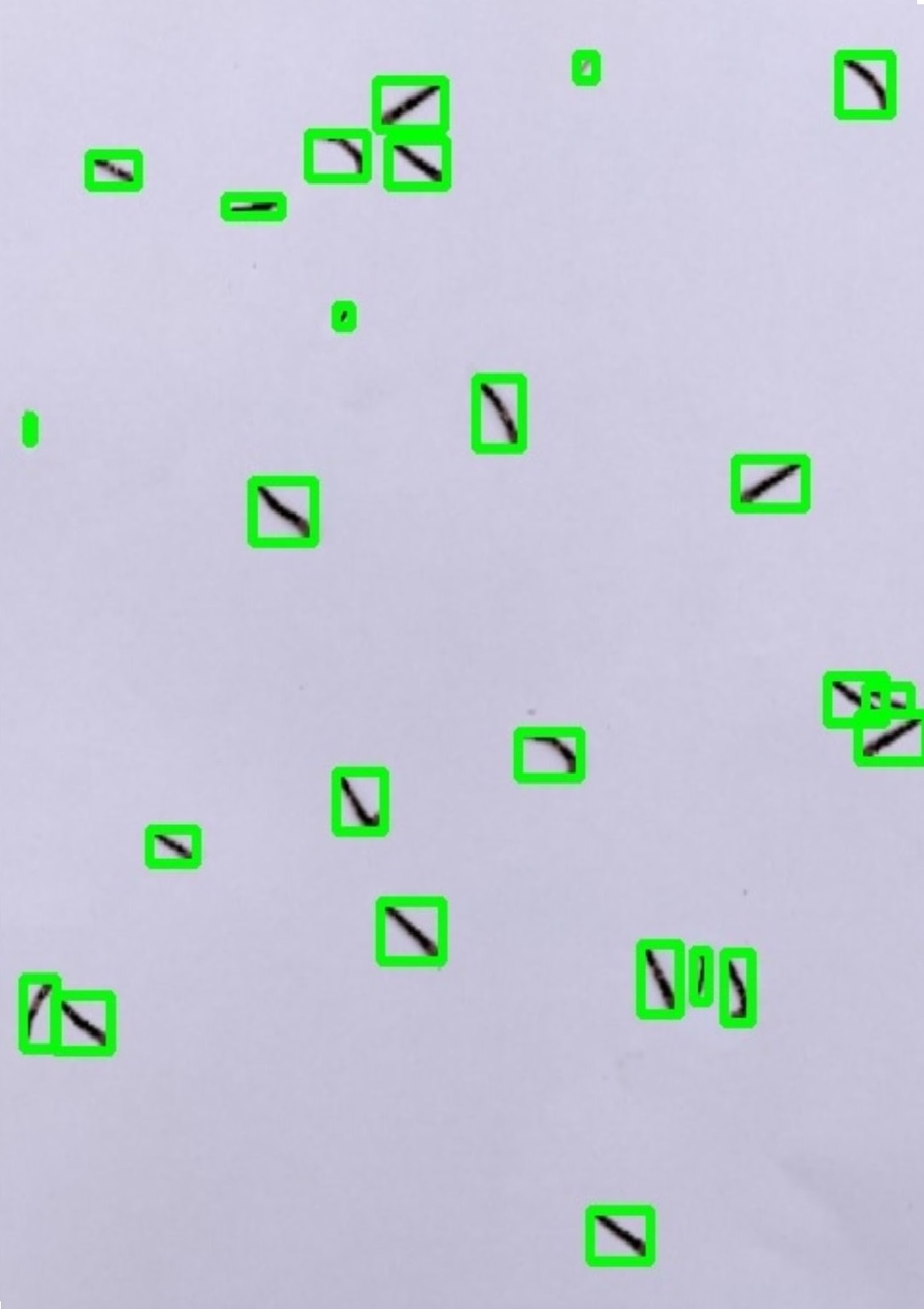}
    \caption{}
    \label{fig:catcount_connected}
  \end{subfigure}
  \caption{Results of calculating caterpillars using image processing: (a) Original image, (b) Thresholding, (c) Erosion, (d) Connected components}
  \label{fig:catcountpaper}
\end{figure}

\begin{figure}[!t]
\centering
\includegraphics[width=\columnwidth]{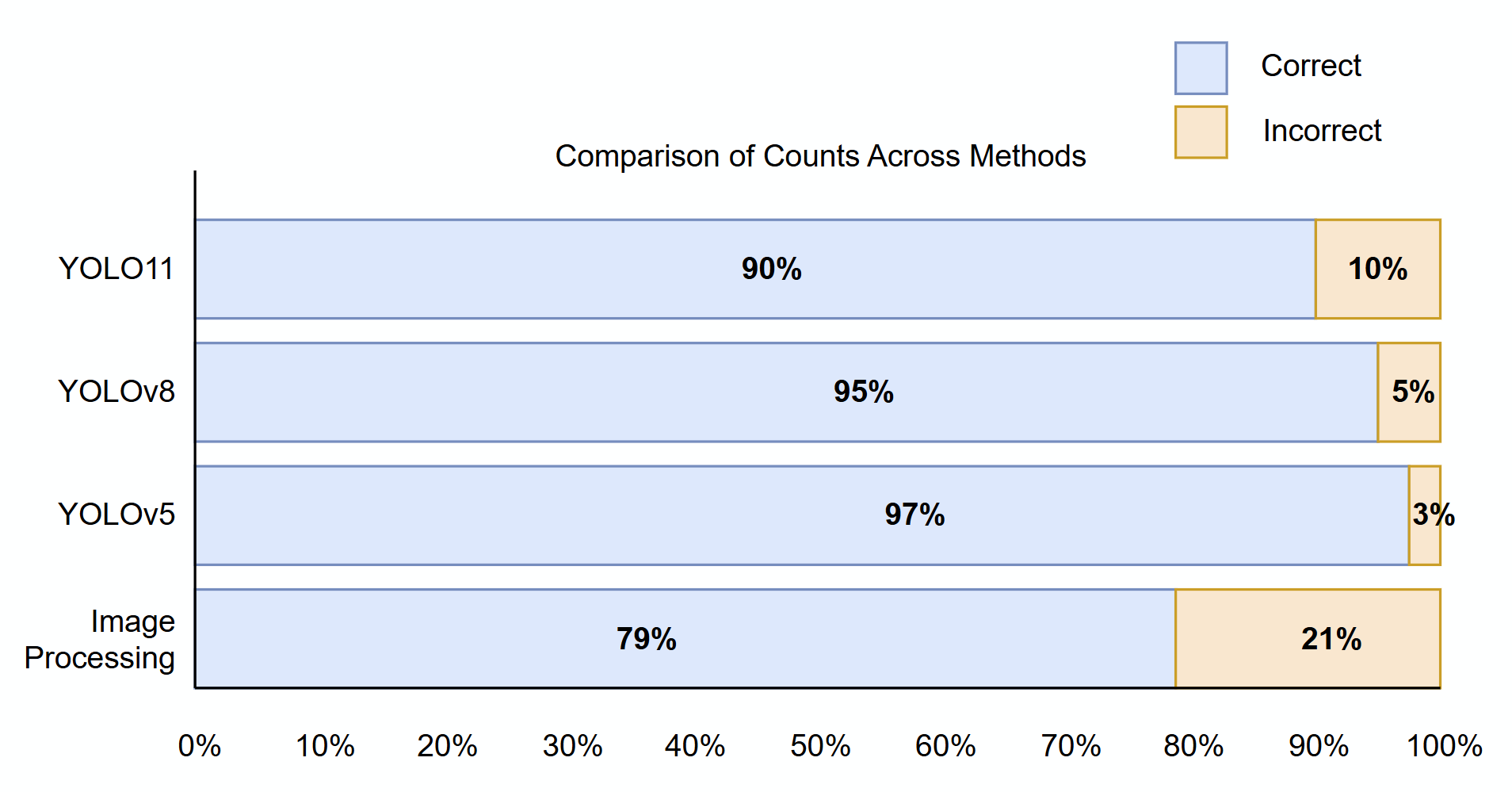}
\caption{Comparison between Image Processing Technique and YOLO (YOLOv5, YOLOv8 \& YOLO11) Object Detection Model on 100 images against ground truth results.}
\label{fig:IPvsYOLO}
\end{figure}

\subsection{Evaluation of CCI identification and classification}

As explained previously, the identification, classification and progression level determination of CCI was achieved using Mask R-CNN and K-means clustering. The models were trained in the above-mentioned environment using 60 validation steps, a detection confidence of 0.9, 100 steps per epoch, and three classes namely, with CCI, without CCI, and background.

The trained model was able to accurately distinguish between infested and non-infested leaflets (healthy and other diseases). Fig. \ref{fig:CCIprocess} illustrates the outcome of the model. Suspicious leaflets (See Fig. \ref{fig:CCIprocess}. (a)) were sent through the Mask-R CNN model for detection. After training the model, as shown in Fig. \ref{fig:CCIprocess}. (b), the results of instance segmentation (masking) of either infested or non-infested leaflets were created. Simultaneously, identification, classification and bounding-box regression were performed to mark the region of leaflets with labels. A rectangular box was drawn around the object after classification. Finally, the FCN layer masks the images according to the coordinates.

After the leaf has been classified and masked, the class label of the image was checked. If the label identifies the image as infected, crop segmentation was performed to separate the mask region (leaflet) from the background (c) as shown in Fig. \ref{fig:CCIprocess}. Only the leaf was extracted while the background was separated using the mask coordination and dimensions. This is a clear example where instance segmentation of Mask RCNN is useful. Following crop segmentation, the HSV range of green colors was converted into a single RGB value (20. 255. 10). Similarly, the brown color caused by caterpillar damage (necrotic regions) was also converted to single RGB color (19. 69. 139). The rest of the colors were turned to white before applying k-means clustering algorithm (see Fig. \ref{fig:CCIprocess}. (d)) to label each pixels within same cluster. The reason for the color conversion is to minimize the number of clusters needed.

Finally, using the 3 clusters of colors (white, green and brown), the number of brown pixels with respect to the leaf area (brown + green) is calculated providing the progression level of the infestation throughout the leaflet. The analysis is more reliable because all pixels are analyzed, including small patches.

For the evaluation of models, performance was assessed by comparing the annotated images with the prediction results during the training process by considering the loss values. After training, metrics such as precision \eqref{eq:precision}, recall \eqref{eq:recall}, Average Precision (AP) \eqref{eq:av_precision} and mean Average Precision (mAP) \eqref{eq:map} value were computed to assess the performance as explained in \cite{HE2022102875} and \cite{Alzubaidi2021}. In object detection, AP is often computed based on the precision-recall curve, which aggregates precision values at different recall levels using \eqref{eq:av_precision}, Where \( P(r_i) \) is the precision at a given recall level \( r_i \).

\begin{equation}
\label{eq:av_precision}
AP = \frac{1}{n} \sum_{i=1}^{n} P(r_i)
\end{equation}

The mAP scores across all classes in the dataset were calculated using \eqref{eq:map}, Where \( C \) is the total number of classes, and \( AP_i \) is the Average Precision for class \( i \).

\begin{equation}
\label{eq:map}
mAP = \frac{1}{C} \sum_{i=1}^{C} AP_i 
\end{equation}

The Mask R-CNN model for CCI identification and classification has performed well in identifying true predictions by achieving precision and recall values of 92\% and 89\% respectively. The respective graphs are shown in Fig. \ref{fig:MaskRCNNGraph}.

Finally, with the use of the above metrics mAP is calculated, and the ResNet101 architecture-based Mask R-CNN model achieved high detection rates with a mAP value of 95.26\%.

\subsection{Evaluation of coconut caterpillar detection and counting}
\subsubsection{Counting caterpillars using object detection (YOLO)}
\label{subsec:YOLOv5}
The YOLO algorithms generated output for both categorization and localization of caterpillars in leaflets. All the YOLO models were trained using hyperparameter configurations on the same computational system to maintain consistency and ensure fair comparability throughout the process. Key training parameters included setting the initial and final learning rates to 0.0001 and 0.01, respectively, ensuring a gradual and controlled learning process. Momentum was fixed at 0.937 to maintain stable updates across epochs, while a minimal weight decay of 0.0005 helped prevent overfitting. The batch size was set at 8, optimizing the balance between memory usage and processing speed. The models were trained on a custom coconut caterpillar dataset, the detection technique was modified to classify two classes (caterpillars and background) and finally, the detection technique was altered to compute the number of caterpillars for each classification. Each model was trained for 100 epochs, providing sufficient time to learn and adapt to the dataset's complexities.

Fig. \ref{fig:catcountImg} shows a sample image of identified caterpillars as well as the result of calculating the number of caterpillars in the image. The comparison among the YOLO modes is given in Figs. \ref{fig:YoloGraph5}, \ref{fig:YoloGraph8}, and \ref{fig:YoloGraph11} for YOLOv5, YOLOv8 and YOLO11 respectively. The summary of the results for each model is shown in Table \ref{table:YOLOPerformance}. As shown in the table, it can be observed that the YOLOv5 clearly outperforms the other two models.

\begin{table}[!t]
\centering
\begin{tabular}{cccc}
\hline
\textbf{Model} & \textbf{Precision (\%)} & \textbf{Recall (\%)} & \textbf{mAP (\%)} \\ \hline
YOLOv5 & 93 & 92.5 & 96.87 \\
YOLOv8 & 90.3 & 92.3 & 96.1 \\
YOLO11 & 91 & 92.2 & 95.9 \\ \hline
\end{tabular}
\caption{Performance metrics for YOLOv5, YOLOv8, and YOLO11.}
\label{table:YOLOPerformance}
\end{table}

\subsubsection{Counting caterpillars using image processing}
\label{subsec:CountCaterpillars}
The automated caterpillar counting process, which simulates the existing manual method of placing caterpillars on paper for counting, is illustrated in Fig. \ref{fig:catcountpaper}. As illustrated in Fig. \ref{fig:catcountpaper}. (a) the captured image was turned into greyscale while using Gaussian Blur to remove background noises to a certain extent. Then thresholding was applied, as shown in Fig. \ref{fig:catcountpaper}. (b), and that helped to clearly distinguish all caterpillars and small dust particles (if available) with the background. To remove the remaining noises (dust particles) erosion was used and the resultant image was given in Fig. \ref{fig:catcountpaper}. (c). This technique was also effective in separating caterpillars that are close or attached. Finally, all the caterpillars are detected by finding the connected components (See Fig. \ref{fig:catcountpaper}. (d)). 

Although this process was fairly accurate, the counting results can be incorrect when the image lighting conditions vary, as this method relies on consistent image processing factors to provide accurate results. Additionally, if the image perception differs, small caterpillars can be removed by the erode function. This research demonstrated that the automated manual process is less accurate compared to the state-of-the-art object detection model mentioned in \ref{subsec:YOLOv5}. 

\section{Discussion}

When comparing the existing studies on coconut disease classification, such as \cite{MARAY2022108399}, \cite{Kadethankar}, and \cite{SINGH2021105986}, they mainly focused on identifying diseases in later stages, when symptoms have progressed visibly. However, it is a fact that identifying diseases in the early stages will give farmers a comparative advantage to apply remedial solutions. Consequently, the proposed method can detect WCLWD with an accuracy of 90\% in the early stages by correctly identifying flaccidity. Flaccidity, which is subtle and often indistinguishable even for experts, marks the initial stages of WCLWD. Hence, the early detection capability of this study addresses a critical gap in current practices and provides significant potential for timely intervention.

Moreover, the proposed method extends beyond disease classification to include severity assessment, an area that has received limited attention in the prior literature. By utilizing inceptionResNetV2 for severity assessment, we achieved an accuracy of 97\%. This dual approach not only enhances the precision of detection but also provides actionable insights into disease progression, enabling more informed decision-making in disease management.

Similarly, traditional and manual methods of counting coconut caterpillars are labor-intensive and prone to inconsistencies, especially under variable conditions such as light conditions and movements of caterpillars. Although, image-processing techniques offer some automation, their accuracy is heavily dependent on the parameters (i.e., lighting conditions, image perception, noise, etc.). Also, limitations such as removal of smaller caterpillars during preprocessing is highly affecting the accuracy of the caterpillar counts.

Our adoption of YOLO-based object detection models significantly improved both accuracy and efficiency of the CCI detection. An analysis was conducted, using 100 images, to compare the accuracies of both YOLO models and the image processing technique against a human counting performed by scientists in CRISL (this was considered as the ground truth). The results obtained are given in Fig. \ref{fig:IPvsYOLO}. As illustrated in the figure, the YOLOv5 model correctly identified the number of caterpillars in 97 of 100 images, leaving 3 images with incorrect values. The YOLOv8 model correctly identified the number of caterpillars in 95 of 100 images, leaving 5 images with discrepancies, while the YOLO11 model accurately identified caterpillars in 90 of 100 images, leaving 10 images with incorrect values. 

Subsequently, among the tested YOLO models, YOLOv5 emerged as the most effective model by achieving a mAP of 96.87\%. Hence, as given in Table \ref{table:YOLOPerformance} and Fig. \ref{fig:IPvsYOLO}, it can be observed that YOLOv5 outperforms both YOLOv8 and YOLO11 in accuracy, precision and recall; YOLOv5 model performance surpasses even the more recent YOLO11 model, underscoring the importance of context-specific model evaluation despite we used the default  \texttt{depth\_multiple} and \texttt{width\_multiple} parameters for all the tested YOLO versions.  

Notably, the image processing technique was capable of accurately identifying only 79 of 100 images. This shows that automated caterpillar detection with YOLO models not only addresses the inefficiencies of manual counting but also ensures reliable results under diverse environmental conditions.

The variety of image resolutions in our dataset played a major role in the success of our models. By using images of different qualities and from different places of the country, we ensured that the models could handle real-world scenarios effectively. This approach allowed our models to perform well even when tested on lower-quality images, highlighting the importance of diverse data during training and testing.

\section{Conclusion}
\label{subsec:Conclusion}
This research study was carried out in Matara, Puttalam, and Makandura in Sri Lanka to find the effectiveness of using Deep Learning techniques such as classification, instance segmentation, and object detection along with image processing techniques for early identification of WCLWD and CCI. The disease classification and the disease severity were determined using CNN, Mask R-CNN, and YOLO models and were able to yield accuracy ranging from 90\% to 97\%. Furthermore, the manual and tedious caterpillar identification process was automated to minimize human inaccuracies and comparative results show that the automated method implemented using YOLOv5 identified caterpillars with only a 3\% error rate. 

In the future, we will work closely with the Coconut Research Institute of Sri Lanka to enhance the early identification of other coconut diseases and pest infestations. We will also explore the use of different architectures for disease detection and severity assessment, to evaluate and compare the performance of models for both WCLWD and CCI. To address real-world applicability, the study will be extended to handle more complex field conditions, such as overlapping leaves, varying orientations (front and back surfaces). Additionally, as a key aspect of future work, Explainable Artificial Intelligence (XAI) techniques will also be incorporated to provide insights into the decision-making process of the models, improving their interpretability. Data from other regions of the country will also be utilized to refine and enhance model accuracy. Finally, the testing will be expanded to other coconut-growing regions, including Kerala, India, to provide sustainable solutions globally. We aim to encourage international researchers to freely use and contribute to the dataset published in \cite{kaggle}, further automating this critical industry.

\section{Acknowledgements}
This research was supported by the Coconut Research Institute of Sri Lanka (CRISL) and the Sri Lanka Institute of Information Technology (SLIIT). The authors would also like to express their sincere gratitude to Dr. Mrs. Nayanie Aratchige, Deputy Director (Research) at the Coconut Research Institute of Sri Lanka, for facilitating the initial field research experiments. Special thanks are extended to Dr. I. M. S. K. Idirisinghe, Mr. P. H. P. Roshan de Silva, Dr. Mrs. Vijitha Vidhanaarachchi, Mr. Anura Pathirana, and Mr. S. P. Manoj for their support in field testing and for providing technical expertise for field sampling. Additionally, we wish to extend our gratitude to R. P. T. I. Gunasekara, P. K. G. C. Akalanka, and H. M. U. D. Rajapaksha for their assistance in conducting the initial experiments and data collection.

\begin{CJK*}{UTF8}{gbsn}
\bibliography{bibliography}

\begin{thebibliography}{10}
\providecommand{\url}[1]{#1}
\csname url@samestyle\endcsname
\providecommand{\newblock}{\relax}
\providecommand{\bibinfo}[2]{#2}
\providecommand{\BIBentrySTDinterwordspacing}{\spaceskip=0pt\relax}
\providecommand{\BIBentryALTinterwordstretchfactor}{4}
\providecommand{\BIBentryALTinterwordspacing}{\spaceskip=\fontdimen2\font plus
\BIBentryALTinterwordstretchfactor\fontdimen3\font minus \fontdimen4\font\relax}
\providecommand{\BIBforeignlanguage}[2]{{%
\expandafter\ifx\csname l@#1\endcsname\relax
\typeout{** WARNING: IEEEtran.bst: No hyphenation pattern has been}%
\typeout{** loaded for the language `#1'. Using the pattern for}%
\typeout{** the default language instead.}%
\else
\language=\csname l@#1\endcsname
\fi
#2}}
\providecommand{\BIBdecl}{\relax}
\BIBdecl

\bibitem{Kumara2015StatusAM}
\BIBentryALTinterwordspacing
A.~D. N.~T. Kumara, M.~Chandrashekharaiah, S.~B. Kandakoor, and A.~K. Chakravarthy, \emph{Status and Management of Three Major Insect Pests of Coconut in the Tropics and Subtropics}.\hskip 1em plus 0.5em minus 0.4em\relax New Delhi: Springer India, 2015, pp. 359--381. [Online]. Available: \url{https://doi.org/10.1007/978-81-322-2089-3_32}
\BIBentrySTDinterwordspacing

\bibitem{Chandy2019}
A.~Chandy, ``Pest infestation identification in coconut trees using deep learning,'' \emph{Journal of Artificial Intelligence and Capsule Networks}, vol.~01, pp. 10--18, 09 2019.

\bibitem{inproceedings}
\BIBentryALTinterwordspacing
T.~Wijesekara, L.~Perera, I.~Wikramananda, I.~Herath, M.~Meegahakumbura, W.~Fernando, and P.~de~Silva, ``Preliminary investigation on weligama coconut leaf wilt disease: A new disease in southern part of sri lanka,'' in \emph{Proceedings of the Second Plantation Crop Symposium}, N.~Nainanayake and J.~Everard, Eds., Colombo, Sri Lanka, 2008, pp. 336--341. [Online]. Available: \url{https://www.researchgate.net/publication/277010023_Preliminary_Investigation_on_Weligama_Coconut_leaf_Wilt_disease_A_new_disease_in_southern_part_of_Sri_Lanka}
\BIBentrySTDinterwordspacing

\bibitem{L.C.P.Fernando}
P.~de~Silva, N.~Aratchige, C.~Ranasinghe, and L.~Perera, ``Diseased palm removals as a strategy for the successful management of weligama coconut leaf wilt phytoplasma disease of coconut in sri lanka,'' \emph{Phytopathogenic Mollicutes}, vol. Vol. 11 (2), pp. 79--85, 03 2022.

\bibitem{L.C.P.Fernando2}
\BIBentryALTinterwordspacing
N.~t. Kumara, L.~Perera, M.~Meegahakumbura, N.~Aratchige, and L.~Fernando, ``Identification of putative vectors of weligama coconut leaf wilt disease in sri lanka,'' pp. 137--146, 04 2015. [Online]. Available: \url{https://www.researchgate.net/publication/276122244_Identification_of_Putative_Vectors_of_Weligama_Coconut_Leaf_Wilt_Disease_in_Sri_Lanka}
\BIBentrySTDinterwordspacing

\bibitem{L.C.P.Fernando3}
L.~Perera, M.~Meegahakumbura, T.~Wijesekara, W.~Fernando, and M.~Dickinson, ``A phytoplasma is associated with the waligama coconut leaf wilt disease in sri lanka.'' \emph{Journal of Plant Pathology}, vol.~94, pp. 205--209, 03 2012.

\bibitem{coconutcaterpillar}
\BIBentryALTinterwordspacing
S.~P. P. J.~S. Company, ``Coconut damaging black-headed caterpillar,'' 2023. [Online]. Available: \url{http://www.spchcmc.vn/EN/Plant-doctor-Detail/Coconut-Damaging-Black-Headed-Caterpillar-5-8392.html}
\BIBentrySTDinterwordspacing

\bibitem{coconutcaterpillar2}
N.~B.~V. Rao, A.~Nischala, G.~Ramanandam, and M.~H.P, ``Biological suppression of coconut black headed caterpillar opisina arenosella outbreak in east godavari district of andhra pradesh-eco friendly technology,'' \emph{Current science}, vol. 115, p. 1588, 11 2018.

\bibitem{CRISL}
\BIBentryALTinterwordspacing
S.~L. Coconut Research~Institute. (2023) Coconut research institue sri lanka. [Online]. Available: \url{https://cri.gov.lk/en/our-team/}
\BIBentrySTDinterwordspacing

\bibitem{WCLWDyellowing}
P.~de~Silva, C.~Perera, B.~Bahder, and R.~Attanayake, ``Nested pcr-based rapid detection of phytoplasma leaf wilt disease of coconut in sri lanka and systemic movement of the pathogen,'' \emph{Pathogens}, vol.~12, 02 2023.

\bibitem{Nainanayaka}
\BIBentryALTinterwordspacing
N.~D, T.~Weerakkody, W.~HTR, P.~Waidyarathne, and S.~WGR, ``Impact of weligama coconut leaf wilt disease (wclwd) on morphological, physiological and yield aspects of coconut palms,'' in \emph{Third symposium on plantation crop research – Stakeholder empowerment through technological advances}, 01 2010. [Online]. Available: \url{https://www.researchgate.net/publication/281445930_Impact_of_Weligama_Coconut_Leaf_Wilt_Disease_WCLWD_on_Morphological_Physiological_and_Yield_Aspects_of_Coconut_Palms}
\BIBentrySTDinterwordspacing

\bibitem{dharmadhikari1977short}
\BIBentryALTinterwordspacing
P.~Dharmadhikari, P.~Perera, and T.~Hassen, ``A short account of the biological control of promecotheca cumingi [col.: Hispidae] the coconut leaf-miner, in sri lanka,'' \emph{Entomophaga}, vol.~22, pp. 3--18, 1977. [Online]. Available: \url{https://doi.org/10.1007/BF02372985}
\BIBentrySTDinterwordspacing

\bibitem{Vidhanaarachchi}
\BIBentryALTinterwordspacing
S.~P. Vidhanaarachchi, P.~Akalanka, R.~Gunasekara, H.~Rajapaksha, N.~Aratchige, D.~Lunugalage, and J.~L. Wijekoon, ``Deep learning-based surveillance system for coconut disease and pest infestation identification,'' in \emph{TENCON 2021 - 2021 IEEE Region 10 Conference (TENCON)}, 2021, pp. 405--410. [Online]. Available: \url{https://doi.org/10.1109/TENCON54134.2021.9707404}
\BIBentrySTDinterwordspacing

\bibitem{presiAgri1}
\BIBentryALTinterwordspacing
A.~Kocian and L.~Incrocci, ``Learning from data to optimize control in precision farming,'' \emph{Stats}, vol.~3, no.~3, pp. 239--245, 2020. [Online]. Available: \url{https://www.mdpi.com/2571-905X/3/3/18}
\BIBentrySTDinterwordspacing

\bibitem{presiAgri2}
\BIBentryALTinterwordspacing
S.~Coulibaly, B.~Kamsu-Foguem, D.~Kamissoko, and D.~Traore, ``Deep learning for precision agriculture: A bibliometric analysis,'' \emph{Intelligent Systems with Applications}, vol.~16, p. 200102, 2022. [Online]. Available: \url{https://www.sciencedirect.com/science/article/pii/S2667305322000400}
\BIBentrySTDinterwordspacing

\bibitem{Elison}
\BIBentryALTinterwordspacing
E.~A. Lins, J.~P.~M. Rodriguez, S.~I. Scoloski, J.~Pivato, M.~B. Lima, J.~M.~C. Fernandes, P.~R.~V. {da Silva Pereira}, D.~Lau, and R.~Rieder, ``A method for counting and classifying aphids using computer vision,'' \emph{Computers and Electronics in Agriculture}, vol. 169, p. 105200, 2020. [Online]. Available: \url{https://www.sciencedirect.com/science/article/pii/S0168169919306039}
\BIBentrySTDinterwordspacing

\bibitem{HanKadipa}
K.~A.~M. Han and U.~Watchareeruetai, ``Classification of nutrient deficiency in black gram using deep convolutional neural networks,'' in \emph{2019 16th International Joint Conference on Computer Science and Software Engineering (JCSSE)}, 2019, pp. 277--282.

\bibitem{YangYu}
\BIBentryALTinterwordspacing
Y.~Yu, K.~Zhang, L.~Yang, and D.~Zhang, ``Fruit detection for strawberry harvesting robot in non-structural environment based on mask-rcnn,'' \emph{Computers and Electronics in Agriculture}, vol. 163, p. 104846, 2019. [Online]. Available: \url{https://www.sciencedirect.com/science/article/pii/S0168169919301103}
\BIBentrySTDinterwordspacing

\bibitem{Francl199757}
\BIBentryALTinterwordspacing
L.~Francl and S.~Panigrahi, ``Artificial neural network models of wheat leaf wetness,'' \emph{Agricultural and Forest Meteorology}, vol.~88, no.~1, pp. 57--65, 1997. [Online]. Available: \url{https://www.sciencedirect.com/science/article/pii/S0168192397000518}
\BIBentrySTDinterwordspacing

\bibitem{Federico}
\BIBentryALTinterwordspacing
F.~Hahn, I.~Lopez, and G.~Hernandez, ``Spectral detection and neural network discrimination of rhizopus stolonifer spores on red tomatoes,'' \emph{Biosystems Engineering}, vol.~89, no.~1, pp. 93--99, 2004. [Online]. Available: \url{https://www.sciencedirect.com/science/article/pii/S1537511004000406}
\BIBentrySTDinterwordspacing

\bibitem{Sladojevic}
S.~Sladojevic, M.~Arsenovic, A.~Anderla, and D.~Stefanović, ``Deep neural networks based recognition of plant diseases by leaf image classification,'' \emph{Computational Intelligence and Neuroscience}, vol. 2016, pp. 1--11, 06 2016.

\bibitem{Wijekoon}
\BIBentryALTinterwordspacing
J.~L. Wijekoon, D.~Nawinna, E.~Gamage, Y.~Samarawickrama, R.~Miriyagalla, D.~Rathnaweera, and L.~Liyanage, \emph{An Intelligent System for Crop Disease Identification and Dispersion Forecasting in Sri Lanka}.\hskip 1em plus 0.5em minus 0.4em\relax Singapore: Springer Singapore, 2022, pp. 187--205. [Online]. Available: \url{https://doi.org/10.1007/978-981-16-9991-7_12}
\BIBentrySTDinterwordspacing

\bibitem{Huang}
K.-Y. Huang, ``Application of artificial neural network for detecting phalaenopsis seedling diseases using color and texture features,'' \emph{Computers and Electronics in Agriculture}, vol.~57, pp. 3--11, 05 2007.

\bibitem{Wang}
Q.~Wang, F.~Qi, M.~Sun, J.~Qu, and J.~Xue, ``Corrigendum to “identification of tomato disease types and detection of infected areas based on deep convolutional neural networks and object detection techniques”,'' \emph{Computational Intelligence and Neuroscience}, vol. 2021, pp. 1--1, 02 2021.

\bibitem{MaskRCNNGeetha}
\BIBentryALTinterwordspacing
B.~Geetha, V.~Jiwatode, R.~Raut, M.~Hussan, M.~Tiwari, and D.~Dobhal, ``An innovative method for detection of insect based on mask-r-cnn approach,'' pp. 559--564, 02 2024. [Online]. Available: \url{https://www.researchgate.net/publication/378122425_An_Innovative_Method_for_Detection_of_Insect_Based_on_Mask-R-CNN_Approach}
\BIBentrySTDinterwordspacing

\bibitem{Manoharan}
\BIBentryALTinterwordspacing
S.~Manoharan, B.~Sariffodeen, K.~Ramasinghe, L.~Rajaratne, D.~Kasthurirathna, and J.~L. Wijekoon, ``Smart plant disorder identification using computer vision technology,'' in \emph{2020 11th IEEE Annual Information Technology, Electronics and Mobile Communication Conference (IEMCON)}, 2020, pp. 0445--0451. [Online]. Available: \url{https://doi.org/10.1109/IEMCON51383.2020.9284919}
\BIBentrySTDinterwordspacing

\bibitem{Hewawitharana}
\BIBentryALTinterwordspacing
G.~Hewawitharana, U.~Nawarathne, A.~Hassan, L.~M. Wijerathna, G.~Sinniah, S.~P. Vidhanaarachchi, J.~Wickramarathne, and J.~L. Wijekoon, ``Effectiveness of using deep learning for blister blight identification in sri lankan tea,'' in \emph{2023 International Research Conference on Smart Computing and Systems Engineering (SCSE)}, vol.~6, 2023, pp. 1--6. [Online]. Available: \url{https://doi.org/10.1109/SCSE59836.2023.10215029}
\BIBentrySTDinterwordspacing

\bibitem{Kasinathan2023}
\BIBentryALTinterwordspacing
T.~Kasinathan and S.~R. Uyyala, ``Detection of fall armyworm (spodoptera frugiperda) in field crops based on mask r-cnn,'' \emph{Signal, Image and Video Processing}, vol.~17, no.~6, pp. 2689--2695, September 2023. [Online]. Available: \url{https://doi.org/10.1007/s11760-023-02485-3}
\BIBentrySTDinterwordspacing

\bibitem{MARAY2022108399}
\BIBentryALTinterwordspacing
M.~Maray, A.~A. Albraikan, S.~S. Alotaibi, R.~Alabdan, M.~A. Duhayyim, W.~K. Al-Azzawi, and A.~alkhayyat, ``Artificial intelligence-enabled coconut tree disease detection and classification model for smart agriculture,'' \emph{Computers and Electrical Engineering}, vol. 104, p. 108399, 2022. [Online]. Available: \url{https://www.sciencedirect.com/science/article/pii/S0045790622006164}
\BIBentrySTDinterwordspacing

\bibitem{Kadethankar}
A.~Kadethankar, N.~Sinha, A.~Burman, and V.~Hegde, ``Deep learning based detection of rhinoceros beetle infestation in coconut trees using drone imagery,'' in \emph{Computer Vision and Image Processing}, S.~K. Singh, P.~Roy, B.~Raman, and P.~Nagabhushan, Eds.\hskip 1em plus 0.5em minus 0.4em\relax Singapore: Springer Singapore, 2021, pp. 463--474.

\bibitem{SINGH2021105986}
\BIBentryALTinterwordspacing
P.~Singh, A.~Verma, and J.~S.~R. Alex, ``Disease and pest infection detection in coconut tree through deep learning techniques,'' \emph{Computers and Electronics in Agriculture}, vol. 182, p. 105986, 2021. [Online]. Available: \url{https://www.sciencedirect.com/science/article/pii/S0168169921000041}
\BIBentrySTDinterwordspacing

\bibitem{Miriyagalla}
\BIBentryALTinterwordspacing
R.~Miriyagalla, Y.~Samarawickrama, D.~Rathnaweera, L.~Liyanage, D.~Kasthurirathna, D.~Nawinna, and J.~L. Wijekoon, ``On the effectiveness of using machine learning and gaussian plume model for plant disease dispersion prediction and simulation,'' in \emph{2019 International Conference on Advancements in Computing (ICAC)}, 2019, pp. 317--322. [Online]. Available: \url{https://doi.org/10.1109/ICAC49085.2019.9103383}
\BIBentrySTDinterwordspacing

\bibitem{kaggle}
S.~Vidhanaarachchi, ``Coconut diseases dataset,'' \url{https://www.kaggle.com/datasets/samitha96/coconutdiseases}, Year of access, accessed: 2023.

\bibitem{iphone6}
\BIBentryALTinterwordspacing
GSMArena, ``Apple iphone 6 - full phone specifications,'' 2024, accessed: 2024-11-17. [Online]. Available: \url{https://www.gsmarena.com/apple_iphone_6-6378.php}
\BIBentrySTDinterwordspacing

\bibitem{iphone11}
{GSMArena}, ``{Apple iPhone 11 - Full phone specifications},'' \url{https://www.gsmarena.com/apple_iphone_11-9848.php}, 2024, accessed: 2024-11-17.

\bibitem{canon}
{Canon Australia}, ``{Canon EOS 3000D - Digital Camera},'' \url{https://www.canon.com.au/cameras/eos-3000d}, 2024, accessed: 2024-11-17.

\bibitem{kumara2015prevalence}
\BIBentryALTinterwordspacing
K.~Kumara, L.~Perera, C.~Ranasinghe, W.~Fernando, R.~Peries, and N.~Adikaram, ``Prevalence of weligama coconut leaf wilt disease in southern sri lanka,'' \emph{Coconut Research Institute of Sri Lanka}, pp. 1--8, 2015. [Online]. Available: \url{https://core.ac.uk/download/pdf/52174498.pdf}
\BIBentrySTDinterwordspacing

\bibitem{coconut_triangle}
{Grow by Coco}, ``The coconut triangle of sri lanka,'' \url{https://growbycoco.com/the-coconut-triangle-of-sri-lanka/}, Year of access, accessed: 2023.

\bibitem{VIA_software}
{Visual Geometry Group, University of Oxford}, ``{VGG Image Annotator (VIA)},'' \url{https://www.robots.ox.ac.uk/~vgg/software/via/}, Year of access, accessed: 2023.

\bibitem{vggref1}
\BIBentryALTinterwordspacing
S.~Bondre and D.~Patil, ``Crop disease identification segmentation algorithm based on mask-rcnn,'' \emph{Agronomy Journal}, vol. 116, no.~3, pp. 1088--1098, 2024. [Online]. Available: \url{https://acsess.onlinelibrary.wiley.com/doi/abs/10.1002/agj2.21387}
\BIBentrySTDinterwordspacing

\bibitem{vggref2}
A.~Gawade, ``Early-stage apple leaf disease prediction using deep learning,'' \emph{Bioscience Biotechnology Research Communications}, vol.~14, pp. 40--43, 03 2021.

\bibitem{Yolov5}
\BIBentryALTinterwordspacing
Ultralytics, ``ultralytics/yolov5: Yolov5 - a state-of-the-art object detection model,'' 2020, accessed: 2024-07-22. [Online]. Available: \url{https://github.com/ultralytics/yolov5}
\BIBentrySTDinterwordspacing

\bibitem{autogyro_yolo_v8_2024}
\BIBentryALTinterwordspacing
Autogyro, ``Yolo-v8,'' 2024, accessed: 2024-12-10. [Online]. Available: \url{https://github.com/autogyro/yolo-V8}
\BIBentrySTDinterwordspacing

\bibitem{ultralytics2024}
\BIBentryALTinterwordspacing
Ultralytics, ``Ultralytics yolo11,'' 2024, accessed: 2024-12-10. [Online]. Available: \url{https://github.com/ultralytics/ultralytics}
\BIBentrySTDinterwordspacing

\bibitem{makesense_ai}
{MakeSense AI}, ``{MakeSense AI},'' \url{https://www.makesense.ai/}, accessed: 2023.

\bibitem{makesenceref1}
L.~Jia, T.~Wang, Y.~Chen, Y.~Zang, X.~Li, H.~Shi, and L.~Gao, ``Mobilenet-ca-yolo: An improved yolov7 based on the mobilenetv3 and attention mechanism for rice pests and diseases detection,'' \emph{Agriculture}, vol.~13, p. 1285, 06 2023.

\bibitem{makesenceref2}
\BIBentryALTinterwordspacing
B.~N. Naik, M.~Ramanathan, and P.~Ponnusamy, ``{Refined single-stage object detection deep-learning technique for chilli leaf disease detection},'' \emph{Journal of Electronic Imaging}, vol.~32, no.~3, p. 033039, 2023. [Online]. Available: \url{https://doi.org/10.1117/1.JEI.32.3.033039}
\BIBentrySTDinterwordspacing

\bibitem{he2017mask}
K.~He, G.~Gkioxari, P.~Dollár, and R.~Girshick, ``Mask {R-CNN},'' in \emph{Proceedings of the {IEEE} International Conference on Computer Vision (ICCV)}, 2017, pp. 2961--2969.

\bibitem{Rahman}
M.~A. Rahman and Y.~Wang, ``Optimizing intersection-over-union in deep neural networks for image segmentation,'' in \emph{Advances in Visual Computing}, G.~Bebis, R.~Boyle, B.~Parvin, D.~Koracin, F.~Porikli, S.~Skaff, A.~Entezari, J.~Min, D.~Iwai, A.~Sadagic, C.~Scheidegger, and T.~Isenberg, Eds.\hskip 1em plus 0.5em minus 0.4em\relax Cham: Springer International Publishing, 2016, pp. 234--244.

\bibitem{Neubeck}
A.~Neubeck and L.~Van~Gool, ``Efficient non-maximum suppression,'' in \emph{18th International Conference on Pattern Recognition (ICPR'06)}, vol.~3, 2006, pp. 850--855.

\bibitem{cropsegmentation}
\BIBentryALTinterwordspacing
S.~Wang, G.~Sun, B.~Zheng, and Y.~Du, ``A crop image segmentation and extraction algorithm based on mask rcnn,'' \emph{Entropy}, vol.~23, no.~9, 2021. [Online]. Available: \url{https://www.mdpi.com/1099-4300/23/9/1160}
\BIBentrySTDinterwordspacing

\bibitem{JAVIDAN2023100081}
\BIBentryALTinterwordspacing
S.~M. Javidan, A.~Banakar, K.~A. Vakilian, and Y.~Ampatzidis, ``Diagnosis of grape leaf diseases using automatic k-means clustering and machine learning,'' \emph{Smart Agricultural Technology}, vol.~3, p. 100081, 2023. [Online]. Available: \url{https://www.sciencedirect.com/science/article/pii/S2772375522000466}
\BIBentrySTDinterwordspacing

\bibitem{bashish2011detection}
D.~Bashish, M.~Braik, and S.~Bani-Ahmad, ``Detection and classification of leaf diseases using k-means based segmentation and neural networks based classification,'' \emph{Information Technology Journal}, 2011.

\bibitem{Mathew2022}
\BIBentryALTinterwordspacing
M.~P. Mathew and T.~Y. Mahesh, ``Leaf-based disease detection in bell pepper plant using yolo v5,'' \emph{Signal, Image and Video Processing}, vol.~16, no.~3, pp. 841--847, 2022. [Online]. Available: \url{https://doi.org/10.1007/s11760-021-02024-y}
\BIBentrySTDinterwordspacing

\bibitem{9153986}
A.~Morbekar, A.~Parihar, and R.~Jadhav, ``Crop disease detection using yolo,'' in \emph{2020 International Conference for Emerging Technology (INCET)}, 2020, pp. 1--5.

\bibitem{yolov8}
\BIBentryALTinterwordspacing
R.~Sapkota and M.~Karkee, ``Comparing yolo11 and yolov8 for instance segmentation of occluded and non-occluded immature green fruits in complex orchard environment,'' 2024. [Online]. Available: \url{https://arxiv.org/abs/2410.19869}
\BIBentrySTDinterwordspacing

\bibitem{nvidia_tesla_t4}
{NVIDIA}, ``{Tesla T4 Tensor Core Product Brief},'' \url{https://www.nvidia.com/content/dam/en-zz/Solutions/Data-Center/tesla-t4/t4-tensor-core-product-brief.pdf}, 2024, accessed: 2024-11-17.

\bibitem{Alzubaidi2021}
L.~Alzubaidi, J.~Zhang, A.~Humaidi \emph{et~al.}, ``Review of deep learning: concepts, cnn architectures, challenges, applications, future directions,'' \emph{Journal of Big Data}, vol.~8, no.~1, p.~53, 2021.

\bibitem{HE2022102875}
\BIBentryALTinterwordspacing
H.~He, H.~Xu, Y.~Zhang, K.~Gao, H.~Li, L.~Ma, and J.~Li, ``Mask r-cnn based automated identification and extraction of oil well sites,'' \emph{International Journal of Applied Earth Observation and Geoinformation}, vol. 112, p. 102875, 2022. [Online]. Available: \url{https://www.sciencedirect.com/science/article/pii/S1569843222000772}
\BIBentrySTDinterwordspacing

\end{thebibliography}
\end{CJK*}

\begin{IEEEbiography}[{\includegraphics[width=1in,height=1.25in,clip,keepaspectratio]{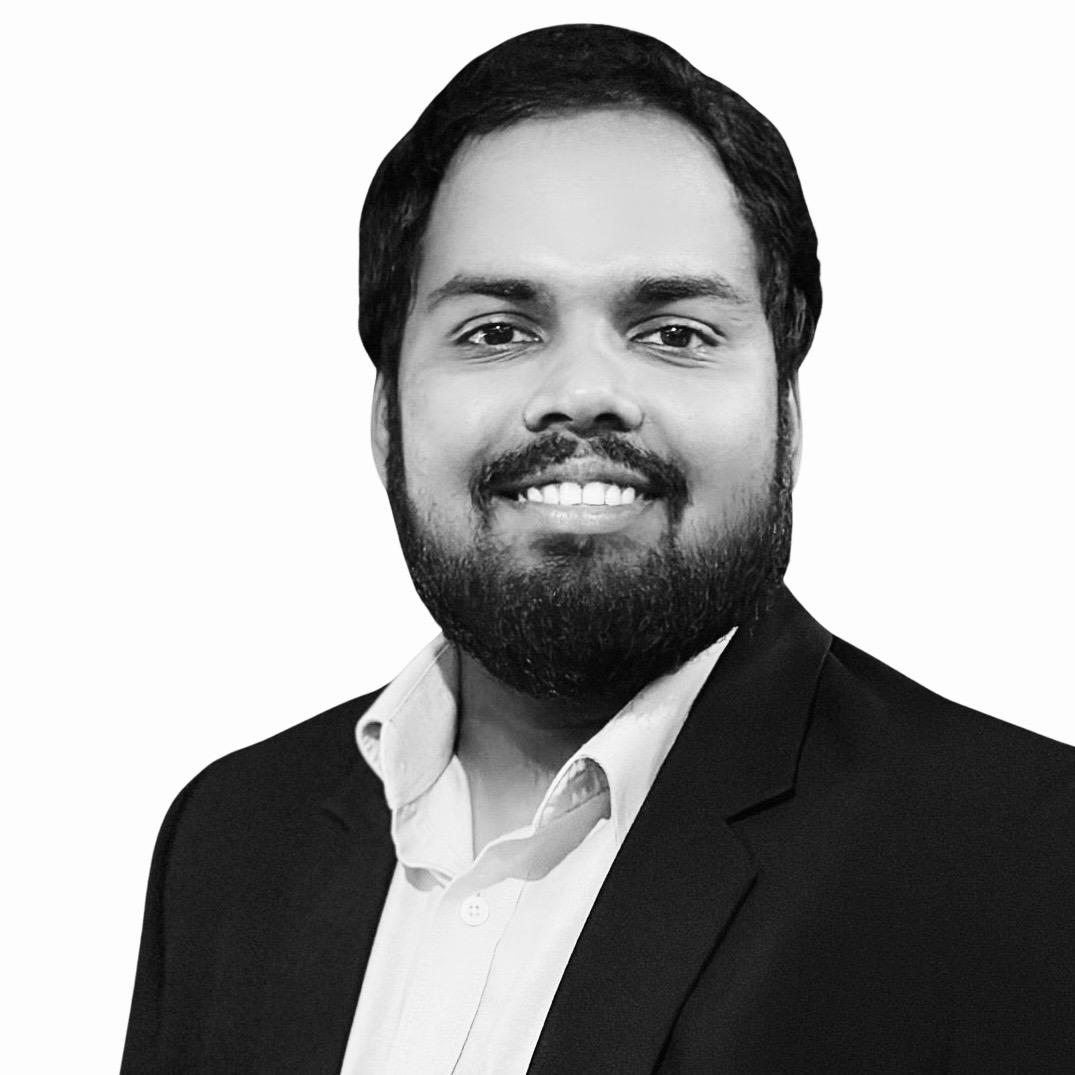}}]{Samitha Vidhanaarachchi} (Member, IEEE) received his B.Sc. degree from the Sri Lanka Institute of Information Technology (SLIIT), Sri Lanka. He has been a lecturer in the Department of Computer Science and Software Engineering at SLIIT. His research interests include developing advanced AI and deep learning solutions in agriculture, enhancing educational tools for diverse needs, and exploring innovative applications across various domains.

\end{IEEEbiography}

\begin{IEEEbiography}[{\includegraphics[width=1in,height=1.25in,clip,keepaspectratio]{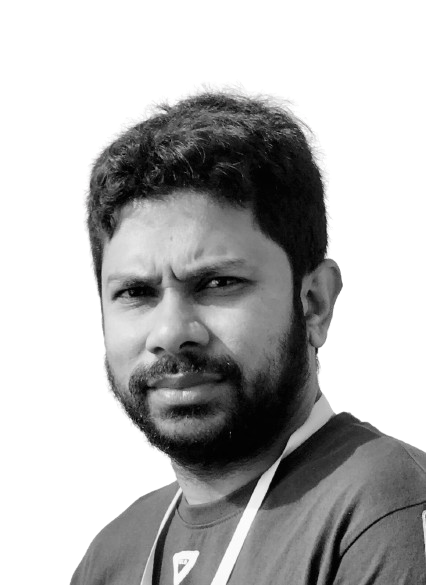}}]{Janaka L. Wijekoon} (Senior Member, IEEE) received his B.Sc. degree from SLIIT, Sri Lanka, in 2010, and went on to complete his M.Sc. (2013) and Ph.D. (2016) degrees at Keio University, Japan. With a background as a postdoctoral researcher with extensive experience in academia, his expertise centres around leveraging AIoT to advance smart societies. His research focuses on collaborative smart infrastructural design and implementations for secure and efficient resource allocation in smart societies and protocol design for data exchange among stakeholders to improve agricultural productivity. 
\end{IEEEbiography}

\begin{IEEEbiography}[{\includegraphics[width=1in,height=1.25in,clip,keepaspectratio]{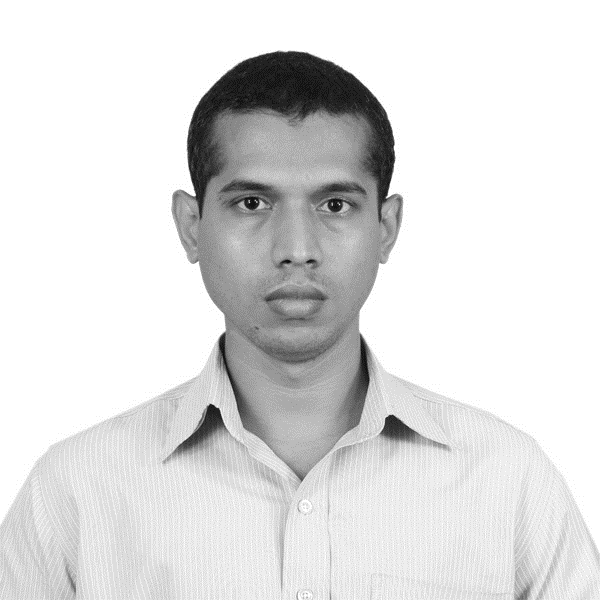}}]{W.A. Shanaka P. Abeysiriwardhana} (Member, IEEE) received his B.Sc. (Hons) and M.Sc. degrees in Electrical Engineering from the University of Moratuwa, Sri Lanka in 2014 and 2015, respectively. He received his PhD degree in Engineering from Keio University, Japan in 2021. His research interests includes intelligent control, network function virtualization, and IoT security.
\end{IEEEbiography}

\begin{IEEEbiography}[{\includegraphics[width=1in,height=1.25in,clip,keepaspectratio]{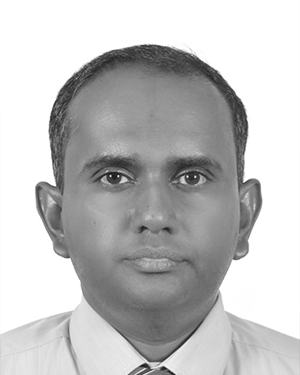}}]{Malitha Wijesundara} (Member, IEEE) earned his BEng degree with Honors in Electronic Engineering from the University of Warwick, United Kingdom in 1998. He then went on to complete his PhD in Computer Engineering at the National University of Singapore. He has extensive experience in Embedded Systems, Wireless Sensor Networks, Media Streaming for E-Learning and Energy Harvesting for Elephant Tracking. His research focuses on developing technology solutions to challenges faced by rural and underprivileged communities. He is a member of IEEE, ISACA and CS(SL).
\end{IEEEbiography}

\EOD
\end{document}